%% file: ema_arxiv_final.tex
\documentclass[10pt,twocolumn,letterpaper]{article}

\usepackage{iccv}
\usepackage{times}
\usepackage{epsfig}
\usepackage{graphicx}
\usepackage{amsmath}
\usepackage{amssymb}

\usepackage{graphicx}
\usepackage{amsmath}
\usepackage{amssymb}
\usepackage{booktabs}
\usepackage{bbding}

\usepackage{pifont}
\usepackage{tabularx}
\usepackage{tablefootnote}
\usepackage[table,xcdraw]{xcolor}

\usepackage{times}
\usepackage{epsfig}
\usepackage{multirow}
\usepackage{silence}
\usepackage{caption}
\usepackage{wrapfig}

\usepackage[breaklinks=true,bookmarks=false]{hyperref}

\iccvfinalcopy % *** Uncomment this line for the final submission

\def\iccvPaperID{****} % *** Enter the ICCV Paper ID here

\ificcvfinal\fi

\begin{document}

%%%%%%%%% TITLE
\title{Efficient Meshy Neural Fields for Animatable Human Avatars}

\author{
Xiaoke Huang\textsuperscript{1}
\quad
Yiji Cheng\textsuperscript{1}
\quad 
Yansong Tang\textsuperscript{1}$^*$
\quad 
Xiu Li\textsuperscript{1}
\quad 
Jie Zhou\textsuperscript{2}
\quad 
Jiwen Lu\textsuperscript{2}
\\[1.5mm]
\{\textsuperscript{1}Shenzhen International Graduate School, \textsuperscript{2}Department of Automation\}, Tsinghua University
}

% \maketitle
\ificcvfinal\thispagestyle{empty}\fi

% \iffalse
\twocolumn[{
    \renewcommand\twocolumn[1][]{#1}
    \maketitle
    \begin{center}
        \includegraphics[width=1.0\linewidth]{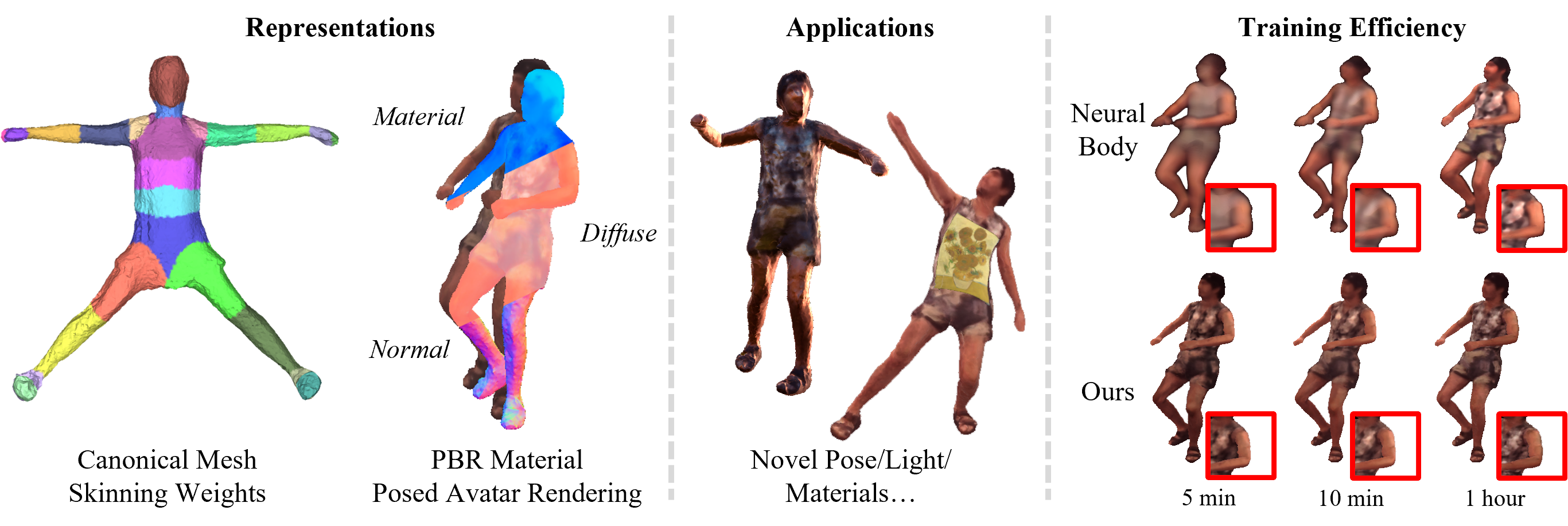}
        \vspace{-1.5em}
        \captionof{figure}{\textbf{EMA} efficiently and jointly learns canonical shapes, materials, and motions via differentiable inverse rendering in an end-to-end manner. The method does not require any predefined templates or riggings. The derived avatars are animatable and can be directly applied to the graphics renderer and downstream tasks. All figures are best viewed in color.}
        \label{fig:teaser}
    \end{center}
}]
% \fi

\let\thefootnote\relax\footnotetext{$^*$Corresponding author.}

%%%%%%%%% ABSTRACT
\input{sections/0-abstract}

%%%%%%%%% BODY TEXT
\input{sections/1-introduction}
\input{sections/2-related_works}
\input{sections/3-method}

\input{sections/4-expreriments}

\input{sections/5-conclusions}

%%%%%%%%% REFERENCES
{\small
\bibliographystyle{ieee_fullname}
\bibliography{ema_bib}
}

\clearpage

\section*{Appendix}
Thank you for reading our supplementary materials!
Here we provide depth descriptions of our method, including details about loss functions (Sec.~\ref{supp:loss_functions}), image-based lighting (Sec.~\ref{supp:light_integration}), and image-based lighting (Sec.~\ref{supp:implementation_details}).
Then we present additional ablations about training views (Sec.~\ref{supp:additional_ablations:num_views}) and skinning module (Sec.~\ref{supp:additional_ablations:skinning}).
Additional experimental results are illustrated in Sec.~\ref{supp:more_comparisons} and Sec.~\ref{supp:mesh_evaluation}.
We showcase application examples in Sec.~\ref{supp:applications}.
In the end, we discuss limitations and social impact in Sec.~\ref{supp:limitation_discussions}.
We strongly encourage our readers to view the supplemental video for a more comprehensive visual perception.

\input{supp/sections/0-loss_functions}
\input{supp/sections/1-light_integration}
\input{supp/sections/2-implementation_details}
\input{supp/sections/3-additional_ablations}
\input{supp/sections/4-more_comparisons}

\input{supp/sections/5-applications}

\input{supp/sections/6-mesh_evaluation}
\input{supp/sections/7-limitations_and_further_discussions}

\end{document}

%% file: sections/0-abstract.tex
\begin{abstract}
\looseness=-1
Efficiently digitizing high-fidelity animatable human avatars from videos is a challenging and active research topic.
Recent volume rendering-based neural representations open a new way for human digitization with their friendly usability and photo-realistic reconstruction quality.
However, they are inefficient for long optimization times and slow inference speed;
their implicit nature results in entangled geometry, materials, and dynamics of humans,
which are hard to edit afterward.
Such drawbacks prevent their direct applicability to downstream applications, especially the prominent rasterization-based graphic ones.
We present \textbf{EMA}, a method that \textbf{E}fficiently learns \textbf{M}eshy neural fields to reconstruct animatable human \textbf{A}vatars. 
It jointly optimizes explicit triangular canonical mesh, spatial-varying material, and motion dynamics, via inverse rendering in an end-to-end fashion.
Each above component is derived from separate neural fields, relaxing the requirement of a template, or rigging.
The mesh representation is highly compatible with the efficient rasterization-based renderer, thus our method only takes about an hour of training and can render in real-time.
Moreover, only minutes of optimization are enough for plausible reconstruction results.
The disentanglement of meshes enables direct downstream applications.
Extensive experiments illustrate the very competitive performance and significant speed boost against previous methods.
We also showcase applications including novel pose synthesis, material editing, and relighting.
The project page: \url{https://xk-huang.github.io/ema/}.
\end{abstract}

%% file: sections/1-introduction.tex
\section{Introduction}

Recent years have witnessed the rise of human digitization~\cite{habermannDeepCapMonocularHuman2020,alexanderCREATINGPHOTOREALDIGITAL,pengNeuralBodyImplicit2021,alldieckDetailedHumanAvatars2018, rajANRArticulatedNeural2020}. This technology greatly impacts the entertainment, education, design, and engineering industry.
There is a well-developed industry solution for this task.
High-fidelity reconstruction of humans can be achieved either with full-body laser scans~\cite{saitoSCANimateWeaklySupervised2021}, dense synchronized multi-view cameras~\cite{xiangModelingClothingSeparate2021a,xiangDressingAvatarsDeep2022a}, or light stages~\cite{alexanderCREATINGPHOTOREALDIGITAL}.
However, these settings are expensive and tedious to deploy and consist of a complex processing pipeline, preventing the technology's democratization.

Another solution is to view the problem as inverse rendering and learn digital humans directly from custom-collected data.
Traditional approaches directly optimize explicit mesh representation~\cite{loperSMPLSkinnedMultiperson2015, fangRMPERegionalMultiperson2018, pavlakosExpressiveBodyCapture2019} which suffers from the problems of smooth geometry and coarse textures~\cite{prokudinSMPLpixNeuralAvatars2020,alldieckVideoBasedReconstruction2018}. Besides, they require professional artists to design human templates, rigging, and unwrapped UV coordinates.
Recently, with the help of volumetric-based implicit representations~\cite{mildenhallNeRFRepresentingScenes2020, parkDeepSDFLearningContinuous2019, meschederOccupancyNetworksLearning2019} and neural rendering~\cite{laineModularPrimitivesHighPerformance2020, liuSoftRasterizerDifferentiable2019, thiesDeferredNeuralRendering2019}, 
one can easily digitize a quality-plausible human avatar from video footage~\cite{jiangNeuManNeuralHuman2022,wengHumanNeRFFreeviewpointRendering}.
Particularly, volumetric-based implicit representations~\cite{mildenhallNeRFRepresentingScenes2020, pengNeuralBodyImplicit2021} can reconstruct scenes or objects with much higher fidelity against previous neural renderer~\cite{thiesDeferredNeuralRendering2019,prokudinSMPLpixNeuralAvatars2020}, and is more user-friendly as it does not need any human templates, pre-set rigging, or UV coordinates.
Captured visual footage and corresponding skeleton tracking are enough for training.
However, better reconstructions and more friendly usability are at the expense of the following factors.
1) \textbf{Inefficiency:}
They require longer optimization times (typically tens of hours or days) and inference slowly.
Volume rendering~\cite{mildenhallNeRFRepresentingScenes2020,lombardiNeuralVolumesLearning2019} formulates images by querying the densities and colors of millions of spatial coordinates. 
In the training stage, due to memory constraints, only a small fraction of points are sampled which leads to slow convergence speed.
2) \textbf{Entangled representations}:
The geometry, materials, and motion dynamics are entangled in the neural networks. 
Due to the implicit nature of neural nets, one can hardly edit one property without touching the others~\cite{yuanNeRFEditingGeometryEditing2022a,liuEditingConditionalRadiance2021}.
3) \textbf{Graphics incompatibility}:
Volume rendering is incompatible with the current popular graphic pipeline,
which renders triangular/quadrilateral meshes efficiently with the rasterization technique.
Many downstream applications require mesh rasterization in their workflow (\eg, editing~\cite{foundationBlenderOrgHome}, simulation~\cite{benderPositionBasedSimulationMethods2015}, real-time rendering~\cite{akenine2019real}, ray-tracing~\cite{waldRTXRayTracing}).
Although there are approaches~\cite{lorensenMarchingCubesHigh,labelleIsosurfaceStuffingFast2007} can convert volumetric fields into meshes, the gaps from discrete sampling degrade the output quality in terms of both meshes and textures.

To address these issues, we present \textbf{EMA}, a method based on \textbf{E}fficient \textbf{M}eshy neural fields to reconstruct animatable human \textbf{A}vatars.
Our method enjoys flexibility from implicit representations and efficiency from explicit meshes, yet still maintains high-fidelity reconstruction quality.
Given video sequences and the corresponding pose tracking, our method digitizes humans in terms of canonical triangular meshes, physically-based rendering (PBR) materials, and skinning weights \textit{w.r.t.} skeletons.
We jointly learn the above components via inverse rendering~\cite{laineModularPrimitivesHighPerformance2020,chenDIBRLearningPredict2021,chenLearningPredict3D2019} in an end-to-end manner.
Each of them is derived from a separate neural field, which relaxes the requirements of a preset human template, rigging, or UV coordinates.
Specifically, we predict a canonical mesh out of a signed distance field (SDF) by differentiable marching tetrahedra~\cite{shenDeepMarchingTetrahedra2021,gaoGET3DGenerativeModel,gaoLearningDeformableTetrahedral2020,munkbergExtractingTriangular3D2022}, then we extend the marching tetrahedra~\cite{shenDeepMarchingTetrahedra2021} for spatial-varying materials by utilizing a neural field to predict PBR materials \textit{on the mesh surfaces} after rasterization~\cite{munkbergExtractingTriangular3D2022,hasselgrenShapeLightMaterial2022,laineModularPrimitivesHighPerformance2020}.
To make the canonical mesh animatable, we take another neural field to model the forward linear blend skinning for the meshes. 
Given a posed skeleton, the canonical mesh is then transformed into the corresponding poses.
Finally, we shade the mesh with a rasterization-based differentiable renderer~\cite{laineModularPrimitivesHighPerformance2020} and train our models with a photo-metric loss.
After training, we export the mesh with materials and discard the neural fields.

\looseness=-1
There are several merits of our method design.
1) \textbf{Efficiency}:
Powered by efficient mesh rendering, our method can render in real-time.
Besides, the training speed is boosted as well, 
since we compute loss holistically on the whole image and the gradients only flow on the mesh surface. In contrast, volume rendering takes limited pixels for loss computation and back-propagates the gradients in the whole space.
Our method only needs about an hour of training and minutes of optimization are enough for plausible avatar reconstruction.
2) \textbf{Disentangled representations}:
Our shape, materials, and motion modules are disentangled naturally by design, which facilitates editing. 
Besides, Canonical meshes with forward skinning modeling handle the out-of-distribution poses better.
3) \textbf{Graphics compatibility}:
Our derived mesh representation is compatible with 
the prominent graphic pipeline, which leads to instant downstream applications (\eg, the shape and materials can be edited directly in design software~\cite{foundationBlenderOrgHome}).
To further improve reconstruction quality, we additionally optimize image-based environment lights and non-rigid motions.

We conduct extensive experiments on standards benchmarks H36M~\cite{ionescuHuman36MLarge2014b} and ZJU-MoCap~\cite{pengNeuralBodyImplicit2021}.
Our method achieves very competitive performance for novel view synthesis, generalizes better for novel poses, 
and significantly improves both training time and inference speed against previous arts.
Our research-oriented code reaches real-time inference speed (100+ FPS for rendering $512\times512$ images).
We in addition showcase applications including novel pose synthesis, material editing, and relighting.

%% file: sections/2-related_works.tex
\section{Related Works}

\begin{figure*}[t]
    \centering
    \includegraphics[width=\textwidth]{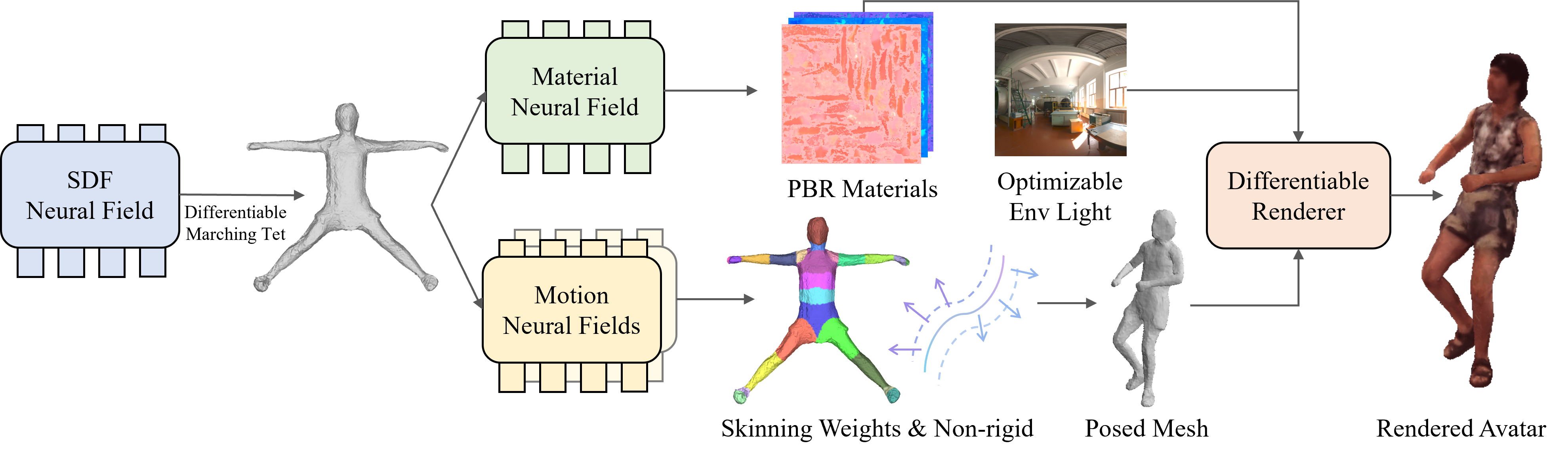}
    \caption{\textbf{The pipeline of EMA}. EMA jointly optimizes canonical shapes, materials, lights, and motions via efficient differentiable inverse rendering. The canonical shapes are attained firstly through the differentiable marching tetrahedra~\cite{gaoLearningDeformableTetrahedral2020,shenDeepMarchingTetrahedra2021,munkbergExtractingTriangular3D2022}, which converts SDF fields into meshes. 
    Next, it queries PBR materials, including diffuse colors, roughness, and specularity on the mesh surface.
    Meanwhile, the skinning weights and per-vertices offsets are predicted on the surface as well, which are then applied to the canonical meshes with the guide of input skeletons. 
    Finally, a rasterization-based differentiable renderer takes in the posed meshes, materials, and environment lights, and renders the final avatars efficiently.}
    \label{fig:my_label}
\end{figure*}

\noindent \textbf{Explicit Representations for Human Modeling:}
It is intuitive to model the surfaces of humans with mesh.
However, humans are highly varied in both shape and appearance and have a large pose space, which all contribute to a high dimensional space.
Researchers first model humans with limited clothes.
One of the prevalent methods is parametric models~\cite{anguelovSCAPEShapeCompletion,loperSMPLSkinnedMultiperson2015,pavlakosExpressiveBodyCapture2019,romeroEmbodiedHandsModeling2022,suANeRFArticulatedNeural2021}.
Fitting humans from scans is inapplicable in real-world applications. 
Thus,~\cite{kanazawaEndtoendRecoveryHuman2018,bogoKeepItSMPL2016,kocabasPAREPartAttention2021,zhangPyMAF3DHuman2021,zhangPyMAFXWellalignedFullbody2022,kocabasVIBEVideoInference2020,sunMonocularOnestageRegression2021} estimate the human surface from images or videos.
To model the clothed human,~\cite{prokudinSMPLpixNeuralAvatars2020,alldieckPhotorealisticMonocular3D2022,alldieckVideoBasedReconstruction2018} deform the template human vertices in canonical T-pose.
However, these methods are prone to capturing coarse geometry due to the limited capacity of the deformation layer.
Besides, the textures are modeled with sphere harmonics which are far from photo-realistic.
Our method takes the mesh as our core representation to enable efficient training and rendering and realize the topological change of shape and photo-realistic texture via neural fields.

\looseness=-1
\noindent \textbf{Implicit Representations for Human Modeling:}
Implicit representations~\cite{parkDeepSDFLearningContinuous2019,meschederOccupancyNetworksLearning2019,mildenhallNeRFRepresentingScenes2020} model the objects in a continuous manner, whose explicit entity cannot be attained directly.
Specifically, Signed Distance Function~\cite{parkDeepSDFLearningContinuous2019}, Occupancy Field~\cite{meschederOccupancyNetworksLearning2019} and Radiance Field~\cite{mildenhallNeRFRepresentingScenes2020} are all parametrized by neural networks.
Given full-body scans as 3D supervision, \cite{saitoPIFuPixelAlignedImplicit2019,saitoPIFuHDMultiLevelPixelAligned2020,heARCHAnimationReadyClothed2022,huangARCHAnimatableReconstruction2020, alldieckPhotorealisticMonocular3D2022} learned the SDFs or occupancy fields directly from images, which could predict photo-realistic human avatars in inference phrase.
\cite{pengNeuralBodyImplicit2021,suANeRFArticulatedNeural2021,liuNeuralActorNeural2022,pengAnimatableNeuralRadiance2021,liTAVATemplatefreeAnimatable2022,jiangNeuManNeuralHuman2022,chenGeometryGuidedProgressiveNeRF2022,wangARAHAnimatableVolume,zhangNDFNeuralDeformable2022,noguchiNeuralArticulatedRadiance2021,zhengStructuredLocalRadiance2022} leveraged the radiance field for more photo-realistic human avatars from multi-view images or single-view videos without any 3D supervision.
Although implicit representations improve reconstruction quality against explicit ones, they still have drawbacks, \eg, large computation burden or poor geometry.s
Besides, volume rendering is incompatible with graphics hardware, 
thus the outputs are inapplicable in downstream applications without further post-processing.
Our method absorbs the merits of implicit representations by using neural networks to predict photo-realistic textures and shape fields, leveraging~\cite{shenDeepMarchingTetrahedra2021} to convert SDFs to explicit meshes, which are fully compatible with the graphic pipeline.

\looseness=-1
\noindent \textbf{Hybrid Representations for Human Modeling:}
There are two tracks of literature modeling humans with explicit geometry representations and implicit texture representations.
One track of literature~\cite{khakhulinRealisticOneshotMeshbased2022,zhaoHighFidelityHumanAvatars2022} leveraged neural rendering techniques~\cite{thiesDeferredNeuralRendering2019}.
Meshes~\cite{prokudinSMPLpixNeuralAvatars2020,zhaoHighFidelityHumanAvatars2022,alldieckVideoBasedReconstruction2018,alldieckDetailedHumanAvatars2018} or point clouds~\cite{wuMultiViewNeuralHuman2020} are commonly chosen explicit representations. Moreover, fine-grained geometry and textures are learned by neural networks.
However, these methods are either only applicable for novel view synthesis~\cite{wuMultiViewNeuralHuman2020} or restricted to self-rotation video captures~\cite{alldieckVideoBasedReconstruction2018,alldieckDetailedHumanAvatars2018}.
Besides, the neural renderers have limitations, \eg, stitching texture~\cite{karrasStyleBasedGeneratorArchitecture2019,karrasAnalyzingImprovingImage2020}, and baked textures into the renderer.
In contrast, the human avatars learned by our method are compatible with graphics pipeline, which means they are \textbf{applicable in downstream tasks}, \eg, re-posing, editing in design software.
The other track of literature leveraged neural networks to learn both geometry and textures based on differentiable rendering~\cite{laineModularPrimitivesHighPerformance2020,chenDIBRLearningPredict2021,chenLearningPredict3D2019}.
It~\cite{liuSoftRasterizerDifferentiable2019,blanzMorphableModelSynthesis,laineModularPrimitivesHighPerformance2020,raviAccelerating3DDeep2020,lassnerPulsarEfficientSpherebased2020,rajANRArticulatedNeural2020} equips traditional graphics pipeline with the ability of error backpropagation, which make scene properties (\textit{i.e,} assets, lights, cameras poses, \textit{etc.}) optimizable through gradient descent \textit{w.r.t} photo-metric loss.
Thus, we can learn both geometry and textures that are compatible with existing graphics hardware.
However, the geometry optimization process is non-convex and highly unstable~\cite{grassalNeuralHeadAvatars2022}, so it is hard to give fine-grained geometry details.
Besides, the topology of the mesh is fixed leading to limited shape modeling.
We leverage~\cite{munkbergExtractingTriangular3D2022,shenDeepMarchingTetrahedra2021} to convert SDFs to meshes with differentiable marching tets, and model the motion dynamics of humans with additional neural fields. 
Our method enjoys flexibility from implicit representations and efficiency brought by explicit meshes, yet still maintains high-fidelity reconstructions.

%% file: sections/3-method.tex
\section{Method}

We formulate the problem as inverse rendering and extend~\cite{munkbergExtractingTriangular3D2022} to model dynamic actors that are driven solely by skeletons.
The canonical shapes, materials, lights, and actor motions are learned jointly in an end-to-end manner.
The rendering happens with an efficient rasterization-based differentiable renderer~\cite{laineModularPrimitivesHighPerformance2020}.

\textbf{Optimization Task}: Let $\Phi$ denote all the trainable parameters (\textit{i.e.,} SDF values and corresponding vertices offset parameters for canonical geometry, spatial-varying and pose-dependent materials and light probe parameters for shading, and forward skinning weights and non-rigid vertices offset parameters for motion).
For a given camera pose $\mathbf{c}$ and a tracked skeleton pose $\mathbf{P}$, 
we render the image $I_{\Phi}(\mathbf{c}, \mathbf{P})$ with a differentiable renderer, and compute loss with a loss function $L$, against the reference image $I_\mathit{ref}(\mathbf{c}, \mathbf{P})$.
The optimization goal is to minimize the empirical risk:
\begin{equation}
    \underset{\phi}{\mathrm{argmin}}\ \mathbb{E}_{\mathbf{c}, \mathbf{P}} \big[L\big( I_{\Phi}(\mathbf{c}, \mathbf{P}), I_\mathit{ref}(\mathbf{c}, \mathbf{P}) \big)\big].
\end{equation}

\noindent The parameters $\Phi$ are optimized with Adam~\cite{kingmaAdamMethodStochastic2017} optimizer.
Following~\cite{munkbergExtractingTriangular3D2022}, our loss function $L = L_\mathrm{img} + L_\mathrm{mask} + L_\mathrm{reg}$ consists of three parts: an image loss $L_\mathit{img}$ using $\ell_1$ norm on tone mapped color, and mask loss $L_\mathrm{mask}$ using squared $\ell_2$, and regularization losses $L_\mathrm{reg}$ to improve the quality of canonical geometry, materials, lights, and motion.
At each optimization step, our method holistically learns both shape and materials from the whole image, while the volume rendering-based implicit counterparts only learn from limited pixels. Powered by an efficient rasterization-based renderer, our method enjoys both faster convergence and real-time rendering speed.
For optimization losses and implementation details, please refer to our supplementary.

\subsection{Canonical Geometry}
\label{method:geometry}

Rasterization-based differentiable renderers take triangular meshes as input, which means the whole optimization process happens over the mesh representation.
Previous works~\cite{alldieckVideoBasedReconstruction2018,alldieckDetailedHumanAvatars2018} require a mesh template to assist optimization as either a good initialization or a shape regularization.
The templates have fixed topology under limited resolutions which harm the geometry quality.
Besides, to make the learned geometry generalize to novel poses, the underlying geometry representations should lie in a canonical space.

We utilize the differentiable marching tetrahedra~\cite{shenDeepMarchingTetrahedra2021,gaoLearningDeformableTetrahedral2020} algorithm, which converts SDF fields into triangular meshes to model the actors in canonical space.
This method enjoys the merit of both template and topology-free from implicit SDF representations and outputs triangle meshes that are directly applicable to rasterization-based renderers.

Let $\mathbf{V}_\mathrm{tet}$, $\mathbf{F}_\mathrm{tet}$, $\mathbf{T}_\mathrm{tet}$ and be the pre-defined vertices, faces, and UV coordinates of the tetrahedra grid.
We parameterize both per-tet vertice SDF value $\mathbf{S}$ and vertices offsets $\Delta \mathbf{V}_\mathrm{tet}$ with a coordiante-based nerual net:
\begin{equation}
    F_{\Phi_\mathit{geom}}: (\mathbf{V}_\mathrm{tet}) \to (\mathbf{S}, \Delta \mathbf{V}_\mathrm{tet}),
\end{equation}

\noindent The canonical mesh $\mathbf{M}_\mathrm{c} = (\mathbf{V}_\mathrm{c} ,\mathbf{F}_\mathrm{c} ,\mathbf{T}_\mathrm{c})$ (\textit{i.e,} canonical mesh vertices, faces, and UV map coordinates) is derived by marching tetrahedra operator $\Pi$:
\begin{equation}
    \Pi: (\mathbf{V}_\mathrm{tet}, \mathbf{F}_\mathrm{tet}, \mathbf{T}_\mathrm{tet}, \mathbf{S}, \Delta \mathbf{V}_\mathrm{tet}) \to \\
    (\mathbf{V}_\mathrm{c} ,\mathbf{F}_\mathrm{c} ,\mathbf{T}_\mathrm{c} ).
\end{equation}

\noindent Specifically, the vertices of the canonical mesh are computed by $\mathbf{v}_c^{ij} = \frac{\mathbf{v}_\mathit{tet}^{\prime i} s_j - \mathbf{v}_\mathit{tet}^{\prime j} s_i}{s_j - s_i}$,
where $\mathbf{v}_\mathit{tet}^{\prime i} = \mathbf{v}^i + \Delta \mathbf{v}^i$ and $\mathrm{sign}(s_i) \neq \mathrm{sign}(s_j)$.
In other words, only the edges that cross the surface of canonical mesh participate the marching tetrahedra operator, which makes the mesh extraction both computation and memory-efficient.

After training, we can discard the SDF and deformation neural nets $F_{\Phi_\mathit{geom}}$ and store the derived meshes. That leads to zero computation overhead in inference time.

\begin{figure*}[t]
\centering
\includegraphics[width=1\linewidth]{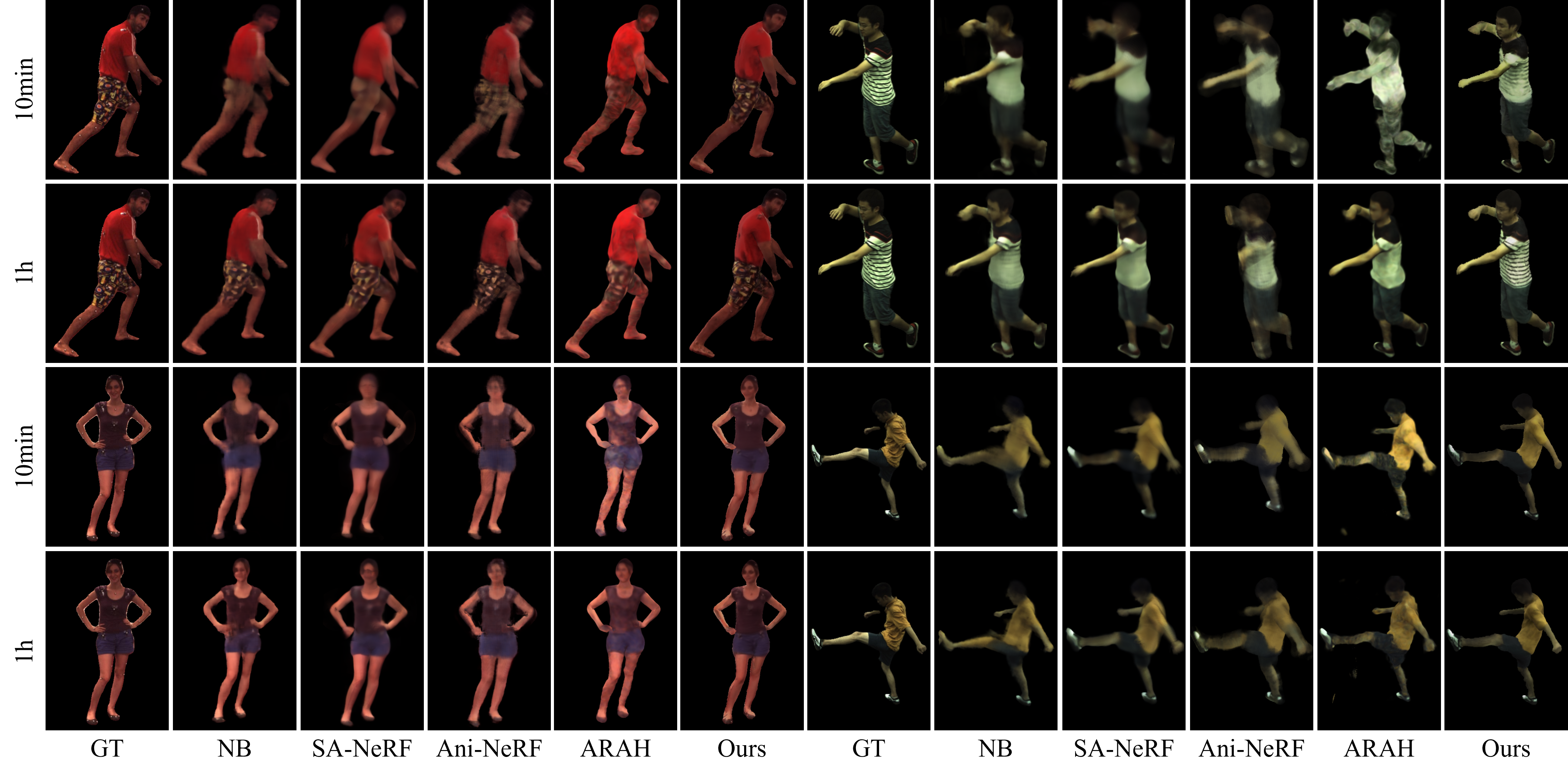}
\vspace{-1em}
\caption{\textbf{Qualitative results of novel view synthesis on the H36M and ZJU-MoCap datasets}. \cite{pengNeuralBodyImplicit2021, pengAnimatableNeuralRadiance2021} generates blurry textures compared with our method. The mesh representations and forward skinning modeling help to improve generalization. Left: Hn36M dataset. Right: ZJU-MoCap dataset. Up: Training pose. Down: Novel pose. \textbf{Zoom in for a better view}.}
\vspace{-1em}
\label{fig:h36m}
\end{figure*}

\subsection{Shading Model}
\label{method:shading}

\textbf{Materials}: we use a physically-based material model~\cite{mcauleyPracticalPhysicallybasedShading2012}, which is directly applicable to our differentiable renderer.
It consists of a diffuse term with an isotopic GGX lobe representing specularity.
Concretely, it consists of three parts: 1) diffuse lobe $\mathbf{k}_d$ has four components, \textit{i.e.} RGB color channels and an additional alpha channel; 
2) specular lobe comprises a roughness value $r$ for GGX normal distribution function 
and a metalness factor $m$ which interpolates the sense of reality from plastic to pure metallic appearance.
The specular highlight color is given by $\mathbf{k}_s = (1-m) \cdot 0.04 + m \cdot \mathbf{k}_d$.
We store the specular lobe into texture $\mathbf{k}_\mathrm{orm} = (o, r, m)$ following the convention, where the channel $o$ is unused.
To compensate for the global illumination, we alternatively store the ambient occlusion value into $o$.
3) normal maps $\mathbf{n}$ represents the fine-grained geometry details.
The diffues color $\mathbf{k}_d$, texture $\mathbf{k}_\mathrm{orm}$, and normal maps $\mathbf{n}$ are parametrized by an neural net:
\begin{equation}
    F_{\Phi_\mathrm{mat}}: (\mathbf{v}_c, \mathbf{P}) \to (\mathbf{k}_d, \mathbf{k}_\mathrm{orm}, \mathbf{n}).
\end{equation}

We query the vertices after rasterization and barycentric interpolation.
The PBR material is further conditioned on poses to model the pose-dependent shading effect.
 
\textbf{Lights}: Our method learns a fixed environment light directly from the reference images. The lights are represented as a cube light.
Given direction $\omega_o$, We follow the render equation~\cite{kajiyaRENDERINGEQUATION1986} to compute the outgoing radiance $L\left(\omega_o\right) $:
\begin{equation}
L\left(\omega_o\right)=\int_{\Omega} L_i\left(\omega_i\right) f\left(\omega_i, \omega_o\right)\left(\omega_i \cdot \mathbf{n}\right) d \omega_i,
\end{equation}

\noindent The outgoing radiance is the integral of the incident radiance $L_i\left(\omega_i\right)$ and the BRDF $f\left(\omega_i, \omega_o\right)$. We do not use spherical Gaussians~\cite{chenLearningPredict3D2019} or spherical harmonics~\cite{bossNeRDNeuralReflectance2021a,zhangPhySGInverseRendering2021} to approximate the image-based lighting.
Instead, we follow~\cite{munkbergExtractingTriangular3D2022} using the split sum approximation that capable of modeling all-frequency image-based lighting:
\begin{equation}
\begin{split}
L\left(\omega_o\right) & \approx \int_{\Omega} f\left(\omega_i, \omega_o\right)\left(\omega_i \cdot \mathbf{n}\right) d \omega_i \\
&\quad \int_{\Omega} L_i\left(\omega_i\right) D\left(\omega_i, \omega_o\right)\left(\omega_i \cdot \mathbf{n}\right) d \omega_i.
\end{split}
\end{equation}

The materials and lights are optimized jointly with geometry and motion modules in an end-to-end manner.
The decomposed design of geometry and shading, along with compatibility with the triangle renderer enables editing and content creation instantly after training. For details about light modeling, please refer to the supplementary.
 
\subsection{Motion Model}
\label{method:motion}

Since we derived mesh-based actors in canonical space with materials and lights, it is intuitive and natural to choose forward linear skinning as our motion model.
Given a skeleton with $B$ bones, the skeleton poses $\mathbf{P} = \{\mathbf{T}_1, \mathbf{T}_2, \ldots, \mathbf{T}_B\}$, where each $\mathbf{T}_i$ represents the transformation on bone $i$, and the blend skinning weights $\mathbf{W} = \{w_1, w_2, \ldots, w_B  \}$, we deform each mesh vertice $\mathbf{v}_c$ in canonical space to the posed vertice $\mathbf{v}_w$ in world space by:
\begin{equation}
    \mathbf{v}_w = \mathrm{LBS}(\mathbf{v}_c, \mathbf{P}, \mathbf{W}) = (\sum\limits_{i=1}^{B} w_i \mathbf{T}_i ) \mathbf{v}_c,
\end{equation}

To compensate for non-rigid cloth dynamics, we further add a layer of pose-dependent non-rigid offsets $\Delta \mathbf{v}_c$ for canonical meshes:
\begin{equation}
    \mathbf{v}_w = \mathrm{LBS}(\mathbf{v}_c  + \Delta \mathbf{v}_c, \mathbf{P}, \mathbf{W}),
\end{equation}

\noindent Where the blend skinning weights and the pose-dependent non-rigid offsets are, respectively, parameterized by neural nets whose inputs are canonical mesh vertices:
\begin{equation}
    F_{\Phi_\mathrm{LBS}}: (\mathbf{v}_c) \to \mathbf{W},
\end{equation}
\begin{equation}
    \quad\; F_{\Phi_\mathrm{nr}}: (\mathbf{v}_c, \mathbf{P}) \to \Delta \mathbf{v}_c.
\end{equation}

Modeling forward skinning is efficient for training as it only forward once in each optimization step, while the volume-based methods~\cite{liTAVATemplatefreeAnimatable2022,wangARAHAnimatableVolume,chenSNARFDifferentiableForward2021} solve the root-finding problem for canonical points in every iteration.
After training, we can export the skinning weight from neural nets which removes the extra computation burden for inference.

%% file: sections/4-expreriments.tex
\input{tables/main-table}

\section{Experiments}

\subsection{Dataset and Metrics}

\noindent \textbf{H36M} consists of 4 multi-view cameras and uses \textbf{marker-based} motion capture to collect human poses.
Each video contains a single subject performing a complex action.
We follow~\cite{pengAnimatableNeuralRadiance2021} data protocol which includes subject S1, S5-S9, and S11.
The videos are split into two parts: ``training poses'' for novel view synthesis and ``Unseen poses'' for novel pose synthesis.
Among the video frames, 3 views are used for training, and the rest views are for evaluation.
The novel view and novel pose metrics are computed on rest views.
We use the same data preprocessing as~\cite{pengAnimatableNeuralRadiance2021}.

\noindent \textbf{ZJU-MoCap} records 9 subjects performing complex actions with 23 cameras.
The human poses are obtained with a markerless motion capture system.
Thus the pose tracking is rather noisier compared with H36M.
Likewise, there are two sets of video frames, ``training poses'' for novel view synthesis and ``Unseen poses'' for novel pose synthesis.
4 evenly distributed camera views are chosen for training, and the rest 19 views are for evaluation.
Again, the evaluation metrics are computed on rest views.
The same data protocol and processing approaches are adopted following~\cite{pengNeuralBodyImplicit2021, pengAnimatableNeuralRadiance2021}.

\noindent \textbf{Metrics}. We follow the typical protocol in ~\cite{pengNeuralBodyImplicit2021, pengAnimatableNeuralRadiance2021} using
two metrics to measure image quality: peak signal-to-noise ratio (PSNR) and structural similarity index (SSIM).

\subsection{Evaluation and Comparison}

\noindent \textbf{Baselines}. We compare our method with template-based methods~\cite{pengNeuralBodyImplicit2021, xuSurfaceAlignedNeuralRadiance2022a} and template-free methods~\cite{pengAnimatableNeuralRadiance2021, wangARAHAnimatableVolume}. Here we list the average metric values with different training times to illustrate our very competitive performance and significant speed boost. 1) Tempelate-based methods. 
Neural Body (NB)~\cite{pengNeuralBodyImplicit2021} learns a set of latent codes anchored to a deformable template mesh to provide geometry guidance.
Surface-Aligned NeRF (SA-NeRF)~\cite{xuSurfaceAlignedNeuralRadiance2022a} proposes projecting a point onto a mesh surface to align surface points and signed height to the surface. 
2) Template-free methods.
Animatable NeRF (Ani-NeRF)~\cite{pengAnimatableNeuralRadiance2021} introduces neural blend weight fields to produce the deformation fields instead of explicit template control.
ARAH~\cite{wangARAHAnimatableVolume} combines an articulated implicit surface representation with volume rendering and proposes a novel joint root-finding algorithm.

\looseness=-1
\noindent \textbf{Comparisons with state-of-the-arts}.
Table~\ref{tab:main-table} illustrates the quantitative comparisons with previous arts.
Notably, our method achieves very competitive performance within much less training time.
The previous volume rendering-based counterparts spend tens of hours of optimization time, while our method only takes an hour of training (for previous SOTA method ARAH~\cite{wangARAHAnimatableVolume}, it takes about 2 days of training).
On the marker-based H36M dataset, our method reaches the SOTA performance in terms of novel view synthesis on training poses and outperforms previous SOTA (ARAH~\cite{wangARAHAnimatableVolume}) for novel view synthesis on novel poses, which indicates that our method can generalize better on novel poses.
The significant boost in training speeds lies in, on the one hand, the core mesh representation which can be rendered efficiently with the current graphic pipeline~\cite{laineModularPrimitivesHighPerformance2020}.
On the other hand, the triangular renderer uses less memory.
Thus we can compute losses over the whole image to learn the representations holistically.
In contrast, previous methods are limited to much fewer sampled pixels in each optimization step.

On the markerless ZJU-Mocap dataset, our method falls behind for training poses novel view synthesis and ranks 3rd place in terms of unseen poses novel view synthesis among the competitors.
We argue that the quality of pose tracking results in the performance gaps between the two datasets.
The markerless pose tracking data \textbf{are much noisier} than the marker-based ones (\eg, the tracked skeleton sequence is jittering, and the naked human~\cite{loperSMPLSkinnedMultiperson2015} rendering is misaligned with human parsings), which makes our performance saturated by harming the multi-view consistency.
The problem is even amplified with the holistic loss computation over the whole pixels.
We conduct additional ablation on pose tracking quality on H36M in Sec.~\ref{exp:data:pose_tracking_type}.
Besides, our non-rigid modeling is only over the surface (no topology change), which is less powerful than the volume rendering-based ones (with topology change).

 We further each method under \textbf{the same optimization duration} in Table~\ref{tab:main-table}.
For the extremely low inference speed of our competitor, we only evaluate at most 10 frames in each subject, and for ZJU MoCap we only choose another 4 evenly spaced cameras as the evaluation views.
For both 1 hour and 10 minutes optimization time, our method outperforms other methods for both training poses and unseen poses novel view synthesis on the marker-based H36M dataset.
On the markerless ZJU-Mocap dataset, our method is comparable with previous SOTA in terms of PSNR and SSIM for both evaluation splits.

\looseness=-1
Figure~\ref{fig:h36m} illustrates the qualitative comparisons between our method and previous arts under the same optimization duration.
It is worth noting that on both H36M and ZJU-MoCap datasets, our method can synthesize clearer and more fine-grained images against competitors, which raises the misalignment of the quantitative metrics for measuring image similarity.

\setlength{\columnsep}{6pt}
\begin{wrapfigure}{r}{0.625\linewidth}
\centering
\vspace{-1em}
\includegraphics[width=1\linewidth]{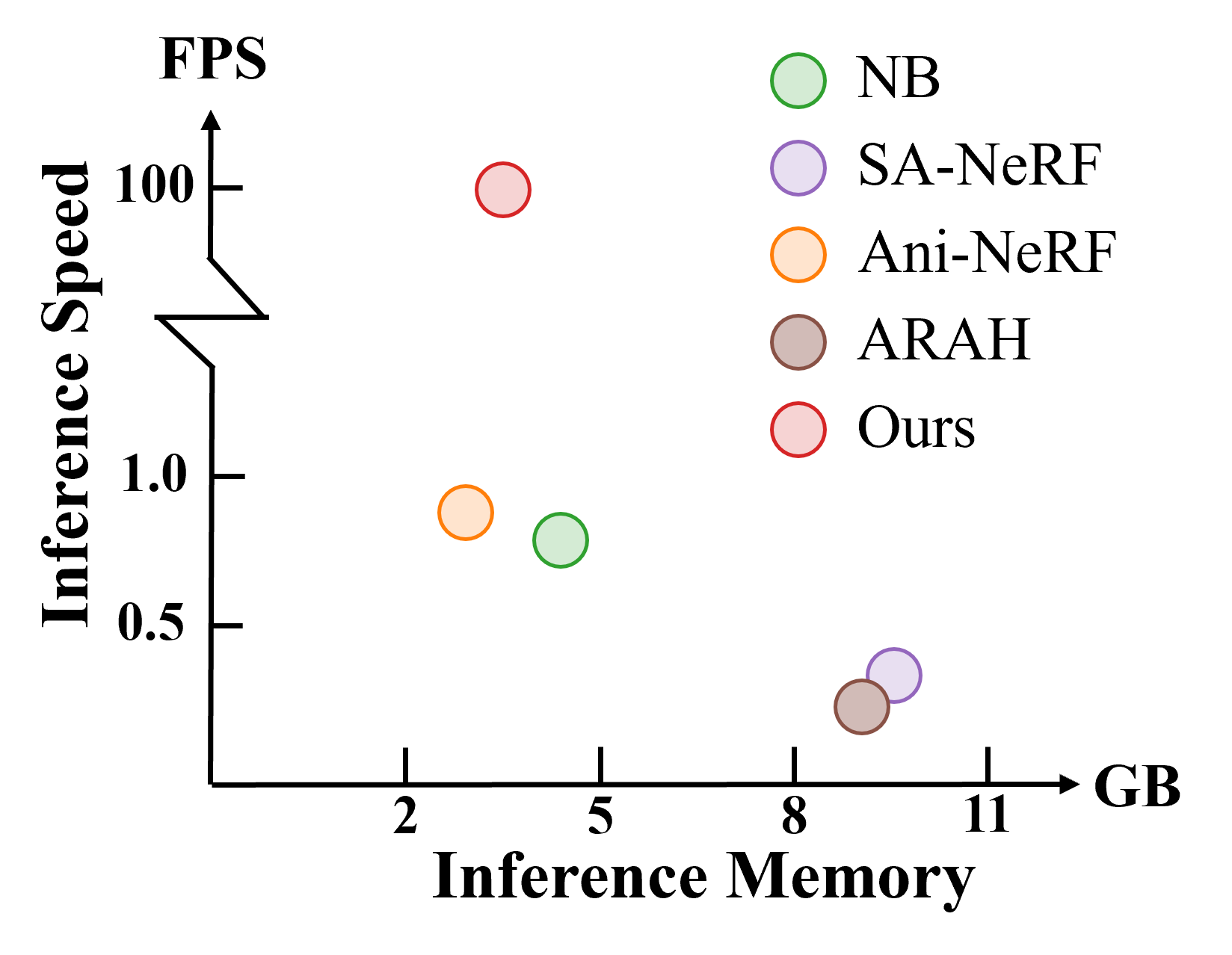}
\vspace{-2em}
\caption{\textbf{Rendering Efficiency}}
\vspace{-1em}
\label{fig:inference}
\end{wrapfigure}

\noindent \textbf{Rendering Efficiency}:
We provide the rendering speed of our method against previous methods in Figure~\ref{fig:inference}. Our method reaches real-time inference speed (100+ FPS for rendering 512×512 images), which is hundreds of times faster than the previous ones. 
Our method takes considerably less memory than the previous ones.

\subsection{Ablation Studies}

\input{tables/ablation-method}

\begin{figure}[t]
\centering
\includegraphics[width=1\linewidth]{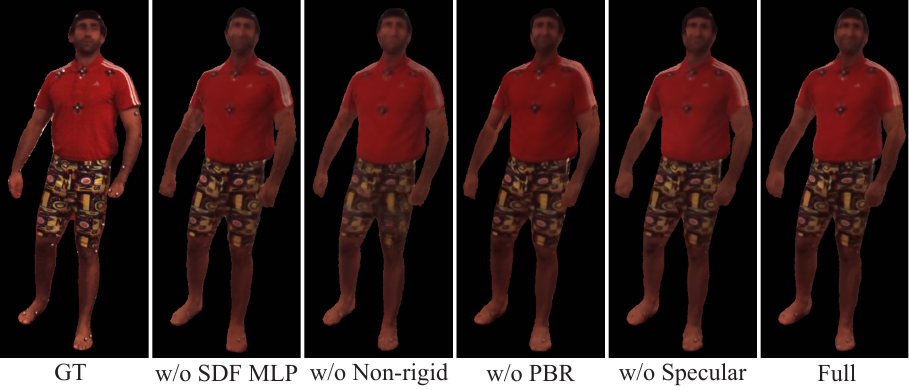}
\caption{\textbf{Qualitative ablation on each module}. The SDF MLP improves the mesh smoothness; non-rigid modeling proves the texture quality by solving the multi-view consistency of cloth dynamics; The PBR materials have a larger capacity for modeling complex materials and lighting against the only-RGB and the no-specular counterparts, which further facilitates both mesh and material learning.}
\label{fig:ablation:method}
\end{figure}

We conduct ablation studies on the H36M S9 subject.

\noindent \textbf{The parametrization type for SDF field.}
The SDF fields can either be parameterized as either MLPs or value fields.
Table~\ref{table:ablation:method} and Figure~\ref{fig:ablation:method} show that using MLP to predict SDF values results in a smoother mesh surface that is watertight.
MLP offers extrapolation ability to predict invisible parts and keep the mesh watertight.
While directly optimizing SDF value fields leads to a jiggling mesh surface and holes in invisible parts during training (\eg, underarm).

\noindent \textbf{The shading model type in geometry module.}
We compare PBR shading models with directly predicting RGB colors and PBR without shading specular.
Table~\ref{table:ablation:method} and Figure~\ref{fig:ablation:method} show that PBR shading models lead to higher metrics against RGB predications,
which indicates that PBR materials can better model complex textures and lights for dynamic humans.
Removing the specular term in PBR does not affect the performance much. 
We conjecture that there is less specularity in human skin and clothes materials.

\noindent \textbf{The impact of the non-rigid net in motion module.}
As shown in Table~\ref{table:ablation:method} and Figures~\ref{fig:ablation:method}, modeling pose-dependent non-rigid dynamics of clothes improves the overall reconstruction quality. It facilitates the aggregation of shading information for multi-view inputs during training.

\begin{figure}[t]
\centering
\includegraphics[width=1\linewidth]{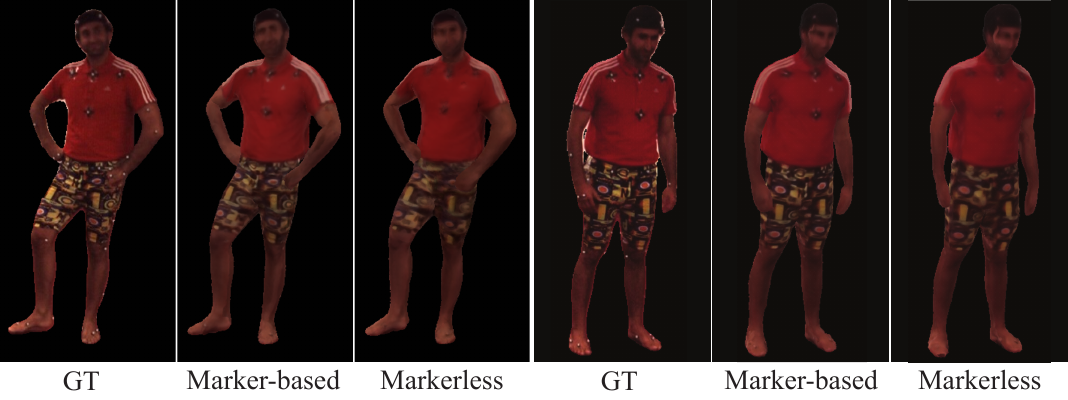}
\vspace{-1em}
\caption{\textbf{Qualitative results of models trained on poses} from marker-less and marker-based systems.}
\label{fig:ablation:data:pose_track_type}
\vspace{-1em}
\end{figure}

\begin{figure}[t]
\centering
\includegraphics[width=1\linewidth]{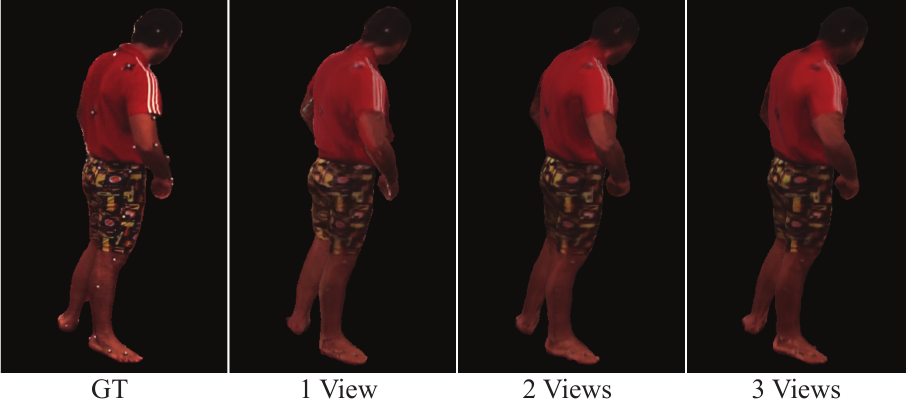}
\vspace{-1em}
\caption{\textbf{Comparison of models trained with different numbers of camera views} on the subject ``S9".}
\label{fig:ablation:data:num_views}
\vspace{-1em}
\end{figure}

\input{tables/ablation-data}

\noindent \textbf{The impact of human tracking quality.}
\label{exp:data:pose_tracking_type}
Table~\ref{table:ablation:data}~(a) and Figure~\ref{fig:ablation:data:pose_track_type} show that using marker-based pose-tracking data can give better results. 
The same phenomenon has been stated in~\cite{pengAnimatableNeuralRadiance2021}.
Noisy marker-less pose-tracking harms the optimization process by damaging the multi-view consistency and the exact pose for shading optimization,
which leads to blurry textures.

\noindent \textbf{The impact of training view amount.}
Table~\ref{table:ablation:data}~(b) and Figure~\ref{fig:ablation:data:num_views} reveal that giving one camera of view degrades the overall reconstruction quality, and multi-view consistency improves the final results.
The model can aggregate multi-view information for better shading optimization, thus leading to clearer surface materials.

\begin{figure}[t]
\centering
\includegraphics[width=1\linewidth]{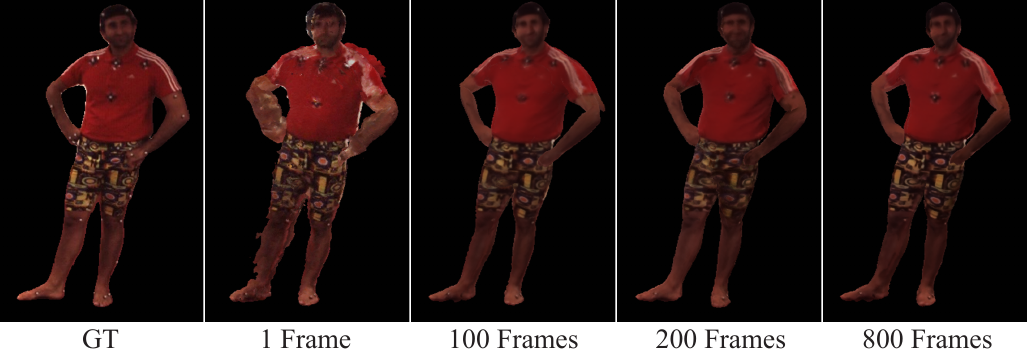}
\vspace{-1em}
\caption{\textbf{Comparison of models trained with different numbers of video frames} on the subject ``S9".}
\label{fig:ablation:data:num_frames}
\vspace{-1em}
\end{figure}

\noindent \textbf{The impact of training frame amount.}
As the number of training frames increases, the rendering quality on novel view and novel pose increases as well (Table~\ref{table:ablation:data}~(c) and Figure~\ref{fig:ablation:data:num_frames}). Notice that the reconstruction quality saturated after using a certain amount of training frames, the same results can be observed in~\cite{pengAnimatableNeuralRadiance2021} as well.

\subsection{Applications}
\label{exp:application}
After training, we can export mesh representations, which enables instant downstream applications.
We showcase two examples of novel pose synthesis, material editing, and human relighting in Figure~\ref{fig:teaser}. For more examples, please refer to our supplementary.

%% file: tables/main-table.tex
\begin{table*}[t]
\centering
\caption{\textbf{Quantitative results}. On the marker-based H36M, our method achieves SOTA performance in all optimization durations. While on the markerless ZJU-MoCap, our method is comparable with previous arts.  ``T.F.'' means template-free; ``Rep.'' means representation; ``T.T'' means the training time; $\ast$ denotes the evaluation on a subset of validation splits.}
\label{tab:main-table}
\resizebox{0.98\textwidth}{!}{%
\begin{tabular}{llll|cccc|cccc}
\toprule
\multicolumn{4}{l|}{}                                                                                                                                & \multicolumn{4}{c|}{H36M}                                                                                                                                                                                                                              & \multicolumn{4}{c}{ZJUMOCAP}                                                                                                                       \\ \hline
\multicolumn{1}{l|}{}                   &                        & \multicolumn{1}{l|}{}                       &                                     & \multicolumn{2}{c}{Training pose}                                                    & \multicolumn{2}{c|}{Novel pose}                                                                             & \multicolumn{2}{c}{Training pose}                             & \multicolumn{2}{c}{Novel pose}                                                    \\ \cline{5-8} \cline{9-12} 
\multicolumn{1}{l|}{\multirow{-2}{*}{}} & \multirow{-2}{*}{T.F.} & \multicolumn{1}{l|}{\multirow{-2}{*}{Rep.}} & \multirow{-2}{*}{T.T.}              & PSNR$\uparrow$                                       & SSIM$\uparrow$                & PSNR$\uparrow$                                       & SSIM$\uparrow$                                       & PSNR$\uparrow$                & SSIM$\uparrow$                & PSNR$\uparrow$                & \multicolumn{1}{c}{SSIM$\uparrow$}                \\ \midrule
\multicolumn{1}{l|}{NB~\cite{pengNeuralBodyImplicit2021}}                 &                        & \multicolumn{1}{l|}{NV}                     & $\sim$10 h                          & 23.31                                                & 0.902                         & {\color[HTML]{1F2329} 22.59}                         & \multicolumn{1}{l|}{{\color[HTML]{1F2329} 0.882}}                         & 28.10                         & 0.944                         & 23.49                         & \multicolumn{1}{l}{0.885}                         \\
\multicolumn{1}{l|}{SA-NeRF~\cite{xuSurfaceAlignedNeuralRadiance2022a}}            &                        & \multicolumn{1}{l|}{NV}                     & $\sim$30 h                          & 24.28                                                & 0.909                         & {\color[HTML]{1F2329} 23.25}                         & \multicolumn{1}{l|}{{\color[HTML]{1F2329} 0.892}}                         & \cellcolor[HTML]{FFFC9E}28.27 & \cellcolor[HTML]{FFFC9E}0.945 & \cellcolor[HTML]{FFFC9E}24.42 & \multicolumn{1}{l}{\cellcolor[HTML]{FFFC9E}0.902} \\
\multicolumn{1}{l|}{Ani-NeRF~\cite{pengAnimatableNeuralRadiance2021}}           & $\checkmark$           & \multicolumn{1}{l|}{NV}                     & $\sim$10 h                          & 23.00                                                & 0.890                         & {\color[HTML]{1F2329} 22.55}                         & \multicolumn{1}{l|}{{\color[HTML]{1F2329} 0.880}}                         & 26.19                         & 0.921                         & 23.38                         & \multicolumn{1}{l}{0.892}                         \\
\multicolumn{1}{l|}{ARAH~\cite{wangARAHAnimatableVolume}}               & $\checkmark$           & \multicolumn{1}{l|}{NV}                     & $\sim$48 h                          & \cellcolor[HTML]{FFCCC9}{\color[HTML]{000000} 24.79} & \cellcolor[HTML]{FFCCC9}0.918 & \cellcolor[HTML]{FFFC9E}{\color[HTML]{1F2329} 23.42} & \multicolumn{1}{l|}{\cellcolor[HTML]{FFFC9E}{\color[HTML]{1F2329} 0.896}} & \cellcolor[HTML]{FFCCC9}28.51 & \cellcolor[HTML]{FFCCC9}0.948 & \cellcolor[HTML]{FFCCC9}24.63 & \multicolumn{1}{l}{\cellcolor[HTML]{FFCCC9}0.911} \\
\multicolumn{1}{l|}{Ours}               & $\checkmark$           & \multicolumn{1}{l|}{Hybr}                   & $\sim$1 h                           & \cellcolor[HTML]{FFFC9E}24.72                        & \cellcolor[HTML]{FFFC9E}0.916 & \cellcolor[HTML]{FFCCC9}23.64                        & \multicolumn{1}{l|}{\cellcolor[HTML]{FFCCC9}0.899}                        & 26.57                         & 0.901                         & 24.38                         & \multicolumn{1}{l}{0.875}                         \\ \midrule
\multicolumn{1}{l|}{NB}                 &                        & \multicolumn{1}{l|}{NV}                     &                                     & 20.58                                                & 0.879                         & 20.27                                                & \multicolumn{1}{l|}{0.867}                                                & \cellcolor[HTML]{FFCCC9}26.87 & \cellcolor[HTML]{FFFC9E}0.922 & 23.67                         & \multicolumn{1}{l}{\cellcolor[HTML]{FFFC9E}0.885}                         \\
\multicolumn{1}{l|}{SA-NeRF}            &                        & \multicolumn{1}{l|}{NV}                     &                                     & 21.03                                                & 0.878                         & 20.71                                                & \multicolumn{1}{l|}{0.869}                                                & 24.92                         & 0.882                         & 23.38                         & \multicolumn{1}{l}{0.869} \\
\multicolumn{1}{l|}{Ani-NeRF}           & $\checkmark$           & \multicolumn{1}{l|}{NV}                     &                                     & 22.54                                                & 0.872                         & 21.79                                                & \multicolumn{1}{l|}{0.856}                                                & 21.23                         & 0.659                         & 20.65                         & \multicolumn{1}{l}{0.652}                         \\
\multicolumn{1}{l|}{ARAH}               & $\checkmark$           & \multicolumn{1}{l|}{NV}                     &                                     & \cellcolor[HTML]{FFFC9E}24.25                        & \cellcolor[HTML]{FFFC9E}0.904 & \cellcolor[HTML]{FFFC9E}23.61                        & \multicolumn{1}{l|}{\cellcolor[HTML]{FFFC9E}0.892}                        & 26.33 & \cellcolor[HTML]{FFCCC9}0.924 & \cellcolor[HTML]{FFCCC9}24.67 & \multicolumn{1}{l}{\cellcolor[HTML]{FFCCC9}0.911} \\
\multicolumn{1}{l|}{Ours}               & $\checkmark$           & \multicolumn{1}{l|}{Hybr}                   & \multirow{-5}{*}{$\sim$1 h$^\ast$}  & \cellcolor[HTML]{FFCCC9}24.83                        & \cellcolor[HTML]{FFCCC9}0.917 & \cellcolor[HTML]{FFCCC9}23.64                        & \multicolumn{1}{l|}{\cellcolor[HTML]{FFCCC9}0.899}                        & \cellcolor[HTML]{FFFC9E}26.66                         & 0.901                         & \cellcolor[HTML]{FFFC9E}24.64 & \multicolumn{1}{l}{0.880} \\ \midrule
\multicolumn{1}{l|}{NB}                 &                        & \multicolumn{1}{l|}{NV}                     &                                     & 20.54                                                & 0.863                         & 20.15                                                & \multicolumn{1}{l|}{0.853}                                                & \cellcolor[HTML]{FFFC9E}25.37 & \cellcolor[HTML]{FFFC9E}0.894                         & 23.54                         & \multicolumn{1}{l}{0.873}                         \\
\multicolumn{1}{l|}{SA-NeRF}            &                        & \multicolumn{1}{l|}{NV}                     &                                     & 20.81                                                & 0.848                         & 20.49                                                & \multicolumn{1}{l|}{0.841}                                                & 24.48                         & 0.878                         & 23.75                         & \multicolumn{1}{l}{0.872}                         \\
\multicolumn{1}{l|}{Ani-NeRF}           & $\checkmark$           & \multicolumn{1}{l|}{NV}                     &                                     & 20.57                                                & 0.822                         & 20.22                                                & \multicolumn{1}{l|}{0.806}                                                & 21.17                         & 0.652                         & 21.16                         & \multicolumn{1}{l}{0.656}                         \\
\multicolumn{1}{l|}{ARAH}               & $\checkmark$           & \multicolumn{1}{l|}{NV}                     &                                     & \cellcolor[HTML]{FFFC9E}23.83                        & \cellcolor[HTML]{FFFC9E}0.895 & \cellcolor[HTML]{FFFC9E}23.13                        & \multicolumn{1}{l|}{\cellcolor[HTML]{FFFC9E}0.884}                        & 25.09 & \cellcolor[HTML]{FFCCC9}0.906 & \cellcolor[HTML]{FFFC9E}24.21 & \multicolumn{1}{l}{\cellcolor[HTML]{FFCCC9}0.898} \\
\multicolumn{1}{l|}{Ours}               & $\checkmark$           & \multicolumn{1}{l|}{Hybr}                   & \multirow{-5}{*}{$\sim$10 m$^\ast$} & \cellcolor[HTML]{FFCCC9}24.27                        & \cellcolor[HTML]{FFCCC9}0.909 & \cellcolor[HTML]{FFCCC9}23.37                        & \multicolumn{1}{l|}{\cellcolor[HTML]{FFCCC9}0.897}                        & \cellcolor[HTML]{FFCCC9}25.51                         & 0.888 & \cellcolor[HTML]{FFCCC9}24.42 & \multicolumn{1}{l}{\cellcolor[HTML]{FFFC9E}0.878} \\
\bottomrule
\end{tabular}%
}
\vspace{-1em}
\end{table*}

%% file: tables/ablation-method.tex
\begin{table}[htbp]
\centering
\caption{
\textbf{The ablation on each module from our method}.
The mesh tends to be noisy and poor for rendering novel poses without MLP parametrization for the geometry module;
Removing the non-rigid module harms the convergence of our model due to the disability to solve multi-view inconsistency;
PBR materials improve the overall shading quality by joint modeling both decomposed materials and lighting.
}
\label{table:ablation:method}
\begin{tabular}{l|cc|cc}
\toprule
              & \multicolumn{2}{c|}{Training Pose} & \multicolumn{2}{c}{Novel Pose}     \\ \hline
              & \multicolumn{1}{c|}{PSNR}  & SSIM  & \multicolumn{1}{c|}{PSNR}  & SSIM  \\ \midrule
w/o SDF MLP   & \multicolumn{1}{c|}{25.17} & 0.913 & \multicolumn{1}{c|}{23.37} & 0.874 \\
w/o Non-rigid & \multicolumn{1}{c|}{25.03} & 0.909 & \multicolumn{1}{c|}{23.45} & 0.877 \\
w/o PBR       & \multicolumn{1}{c|}{25.10} & 0.914 & \multicolumn{1}{c|}{23.44} & 0.878 \\
w/o Specular  & \multicolumn{1}{c|}{25.24} & 0.915 & \multicolumn{1}{c|}{\textbf{23.58}} & \textbf{0.879} \\ \midrule
Full          & \multicolumn{1}{c|}{\textbf{25.26}} & \textbf{0.916} & \multicolumn{1}{c|}{23.52} & \textbf{0.879} \\
\bottomrule
\end{tabular}
\end{table}

%% file: tables/ablation-data.tex
\begin{table}[htbp]
\centering
\caption{\textbf{The ablations results on data quality and quantity} on H36M S9 subject,  in terms of PSNR and SSIM (higher is better). The better the data quality, the better the reconstruction results.}
\label{table:ablation:data}
\begin{tabular}{ccccc}
\toprule
\multicolumn{1}{l|}{}    & \multicolumn{2}{c|}{Training pose}                         & \multicolumn{2}{c}{Novel pose}     \\ \hline
\multicolumn{1}{l|}{}              & \multicolumn{1}{c|}{PSNR$\uparrow$}  & \multicolumn{1}{c|}{SSIM$\uparrow$}  & \multicolumn{1}{c|}{PSNR$\uparrow$}  & \multicolumn{1}{c}{SSIM$\uparrow$} \\ \midrule
\multicolumn{5}{l}{\textbf{(a)} type of pose tracking}                                                                                     \\ \midrule
\multicolumn{1}{l|}{w/o marker}  & \multicolumn{1}{l|}{24.73} & \multicolumn{1}{l|}{0.893} & \multicolumn{1}{l|}{22.60} & 0.853                    \\
\multicolumn{1}{l|}{w/ marker} & \multicolumn{1}{l|}{\textbf{25.53}} & \multicolumn{1}{l|}{\textbf{0.911}} & \multicolumn{1}{l|}{\textbf{23.80}} & \textbf{0.879}                    \\ \midrule
\multicolumn{5}{l}{\textbf{(b)} number of training views}                                                                                     \\ \midrule
\multicolumn{1}{l|}{1 view}   & \multicolumn{1}{l|}{25.09} & \multicolumn{1}{l|}{0.906} & \multicolumn{1}{l|}{22.97} & 0.866 \\
\multicolumn{1}{l|}{2 views}   & \multicolumn{1}{l|}{25.56} & \multicolumn{1}{l|}{\textbf{0.911}} & \multicolumn{1}{l|}{\textbf{23.76}} & \textbf{0.878} \\
\multicolumn{1}{l|}{3 views}   & \multicolumn{1}{l|}{\textbf{25.57}} & \multicolumn{1}{l|}{\textbf{0.911}} & \multicolumn{1}{l|}{23.67} & 0.876 \\ \midrule
\multicolumn{5}{l}{\textbf{(c)} number of training frames}                                                                                    \\ \midrule
\multicolumn{1}{l|}{1 frame}   & \multicolumn{1}{l|}{20.93} & \multicolumn{1}{l|}{0.817} & \multicolumn{1}{l|}{19.58} & 0.785 \\
\multicolumn{1}{l|}{100 frames} & \multicolumn{1}{l|}{23.99} & \multicolumn{1}{l|}{0.882} & \multicolumn{1}{l|}{22.49} & 0.856 \\
\multicolumn{1}{l|}{200 frames} & \multicolumn{1}{l|}{\textbf{25.27}} & \multicolumn{1}{l|}{\textbf{0.905}} & \multicolumn{1}{l|}{\textbf{23.32}} & \textbf{0.873} \\
\multicolumn{1}{l|}{800 frames} & \multicolumn{1}{l|}{24.89} & \multicolumn{1}{l|}{0.900} & \multicolumn{1}{l|}{23.16} & \textbf{0.873} \\ 
\bottomrule
\end{tabular}
\end{table}

%% file: sections/5-conclusions.tex
\section{Discussions and Conclusions}

\noindent 
\textbf{Discussions}:
Our method leverage mesh as our core representation, which enables us efficiency for both training and rendering.
However, the resolution of mesh is fixed in our pipeline, preventing fine-grained geometry and texture reconstruction.
One possible solution could be tetrahedra grids sub-division~\cite{schaeferSmoothSubdivisionTetrahedral,gaoTetGANConvolutionalNeural2022,kalischekTetrahedralDiffusionModels2022a}.
But it may break the SDF values around the derived meshes 
since there is no regularization over the whole SDF field.
Our non-rigid modeling has less capacity, since we assume there is no topology change of mesh \textit{wrt.} the non-rigid motion.
Otherwise, we cannot query materials and motions in the canonical shape.
One can solve it via the dense correspondence between the meshes before and after applying non-rigid motions~\cite{ahmedDenseCorrespondenceFinding2008,zeng3DHumanMesh2021},
yet such an operation may increase computation drastically.

\noindent \textbf{Conclusions}: we present EMA, which learns human avatars through hybrid meshy neural fields efficiently.
EMA jointly learns hybrid canonical geometry, materials, lights, and motions via a rasterization-based differentiable renderer. 
It only requires one hour of training and can render in real-time with a triangle renderer.
Minutes of training can produce plausible results.
Our method enjoys flexibility from implicit representations and efficiency from explicit meshes.
Experiments on the standard benchmark indicate the competitive performance and generalization results of our method.
The digitized avatars can be directly used in downstream tasks.
We showcase examples including novel pose synthesis, material editing, and human relighting.

%% file: supp/sections/0-loss_functions.tex
\section{Loss Functions}

\label{supp:loss_functions}

Our loss function $L = L_\mathrm{img} + L_\mathrm{mask} + L_\mathrm{reg}$ is composed of three parts: an image loss $L_\mathit{img}$ using $\ell_1$ norm on tone mapped colors, and mask loss $L_\mathrm{mask}$ using squared $\ell_2$, and regularization losses $L_\mathrm{reg}$ to improve the quality of canonical geometry, materials, lights, and motion.

\noindent \textbf{Image loss}: 
our renderer utilizes physically-based shading to produce high-dynamic range (HDR) images.
Then the complex materials and environmental lights are elaborately optimized.
Thus our loss function requires a full range of floating point values.
We follow~\cite{hasselgrenAppearanceDrivenAutomatic3D2021,munkbergExtractingTriangular3D2022,hasselgrenShapeLightMaterial2022} to compute $\ell_1$ norm on tone mapped colors.
Specifically, we first transform linear radiance values $i$ according to a tone-mapping operator $T(i)=\Gamma(\log (i+1))$, in which $\Gamma(i)$ is a linear RGB to sRGB transformation function~\cite{stokesStandardDefaultColor1996}:
\begin{equation}
\begin{aligned}
\Gamma(i) & = \begin{cases}12.92 i & i \leq 0.0031308 \\
(1+a) i^{1 / 2.4}-a & i>0.0031308\end{cases} \\
a & =0.055,
\end{aligned}
\end{equation}

\noindent \textbf{Mask loss}:
The renderer~\cite{laineModularPrimitivesHighPerformance2020} renders both the shaded images and the corresponding rasterization masks in a differentiable manner.
Therefore, we compute the $\ell_2$ norm between the masks and the preprocessed mattings (in both ZJU-MoCap and H36M benchmarks, we use the provided preprocessed subject masks from~\cite{pengNeuralBodyImplicit2021,gongInstanceLevelHumanParsing2018}), akin to the traditional shape-from-silhouette~\cite{maInvitation3DVision2004} technique.
The mask loss is parallel with the image loss, yet facilitates the course of shading optimization by making shape convergence super fast in about a hundred training steps.

\noindent \textbf{Regularizers}:
We need various priors to encourage the optimization to converge at a place where the geometry, materials, and lighting are well separated and smooth enough~\cite{munkbergExtractingTriangular3D2022,hasselgrenShapeLightMaterial2022}.
Therefore, we choose to minimize regularization during training.

\noindent We introduce smoothness to PBR materials in terms of albedo $\mathbf{k}_d$, specular parameters $\mathbf{k}_\mathrm{orm}$, and surface geometry nomral $\mathbf{n}$ as following:
\begin{equation}
L_{\mathbf{k}}=\frac{1}{\left|\mathbf{x}_{\text {surf }}\right|} \sum_{\mathbf{x}_{\text {surf }}}\left|\mathbf{k}\left(\mathbf{x}_{\text {surf }}\right)-\mathbf{k}\left(\mathbf{x}_{\text {surf }}+\mathbf{\epsilon}\right)\right|,
\end{equation}

\noindent where ${\left|\mathbf{x}_{\text {surf }}\right|}$ is a surface point on the surface in canonical space and $\mathbf{\epsilon} \sim \mathcal{N}(0,\sigma\!\!=\!\!0.01)$ is a small random offset.
We regularize the geometry normal on the surface of the canonical mesh derived from the SDF field for a seek of a smoother surface and avoidance of holes in the surface.

\noindent We regularize light by assuming the neutral spectrum in the real world.
Specifically, given the per-channel average radiance densities $\Bar{c_i}$, we penalize the color shifts as:
\begin{equation}
L_{\text {light }}=\frac{1}{3} \sum_{i=0}^3 \left| {c_i}-\frac{1}{3} \sum_{i=0}^3 {c_i} \right|,
\end{equation}

To encourage a watertight surface and reduce floating meshes both inside and outside the subject, 
we impose regularizations on the SDF field as:
\begin{equation}
\begin{aligned}
L_{\mathrm{sdf}}=\sum_{i, j \in \mathrm{S}_e} &H\left(\sigma\left(s_i\right), \operatorname{sign}\left(s_j\right)\right) \\
& +H\left(\sigma\left(s_j\right), \operatorname{sign}\left(s_i\right)\right),
\end{aligned}
\end{equation}

\noindent where $\mathrm{S}_e$ is the set of all vertex along their edges in which the signs of the SDF values are different (\ie, $\mathrm{sign}(s_i) \neq \mathrm{sign}(s_j)$).
To remove the floating meshes outside the surface, we impose an additional loss.
For a triangle surface $f$ extracted by marching tetrahedra, if $f$ is invisible, we encourage its SDF values to be positive as:
\begin{equation}
L_{\mathrm{invis}}=\sum_{i \in \mathrm{S}_{\mathrm{invis}}} H(\sigma\left(s_i\right), 1).
\end{equation}

We weigh the above terms and use the loss for all our experiments:
\begin{equation}
\begin{aligned}
	L &= L_\text{image} + L_\text{mask} \\
	&+ \underbrace{\lambda_{\mathbf{k}_{\mathrm{d}}}}_{=0.03} L_{\mathbf{k}_{\mathrm{d}}} 
	+ \underbrace{\lambda_{\mathbf{k}_{\mathrm{orm}}}}_{=0.05} L_{\mathbf{k}_{\mathrm{orm}}}
	+ \underbrace{\lambda_{\mathbf{n}}}_{=0.025} L_{\mathbf{n}} \\
	&+ \underbrace{\lambda_\text{light}}_{=0.005} L_\text{light}
 	+ \underbrace{\lambda_\text{sdf}}_{=0.02} L_\text{sdf}
  	+ \underbrace{\lambda_\text{invis}}_{=0.01} L_\text{invis}.
\end{aligned}
\end{equation}

%% file: supp/sections/1-light_integration.tex
\section{Image-based Lighting}

\label{supp:light_integration}
The split sum shading model is widely used in real-time rendering~\cite{akenine2019real}, giving both realism and efficiency against spherical Gaussians (SG) and spherical harmonics (SH)~\cite{bossNeRDNeuralReflectance2021a,chenLearningPredict3D2019,zhangPhySGInverseRendering2021}.
We use a differentiable split sum~\cite{karisRealShadingUnreal2013} shading model to approximate rendering equation~\cite{kajiyaRENDERINGEQUATION1986} for image-based environment light learning as~\cite{munkbergExtractingTriangular3D2022}:
\begin{equation}
\begin{split}
L\left(\omega_o\right) & \approx \int_{\Omega} f\left(\omega_i, \omega_o\right)\left(\omega_i \cdot \mathbf{n}\right) d \omega_i \\
&\quad \int_{\Omega} L_i\left(\omega_i\right) D\left(\omega_i, \omega_o\right)\left(\omega_i \cdot \mathbf{n}\right) d \omega_i.
\label{supp:euq:render}
\end{split}
\end{equation}

\noindent where $D$ is the GGX normal distribution function (NDF)~\cite{walterMicrofacetModelsRefraction} in a Cook-Torrance microfacet specular shading model~\cite{cookReflectanceModelComputer}.
The first term contributes to the specular BSDF \textit{wrt.} a solid white environment light, which depends solely on the roughness $r$ of the BSDF and the light-surface angles $\cos \theta = \omega_i \cdot \mathbf{n}$.
The second term contributes to the integral of the incoming radiance with the GGX normal distribution function, $D$.
Both terms can be pre-computed and filtered to reduce computation~\cite{karisRealShadingUnreal2013}.

The training parameters are texels of a cube light map whose resolution is $6 \times 512 \times 512$.
The pre-integrated lighting for the least roughness values is derived from the base level, and multiple smaller mip levels are constructed from it~\cite{karisRealShadingUnreal2013}.
Each mip-map is filtered by average-pooling the base level of the current resolution and is convolved with the GGX normal distribution function.
The per mip-level filter bounds are pre-computed as well.
We leverage a PyTorch implementation with CUDA extensions from~\cite{munkbergExtractingTriangular3D2022}.
Moreover, a cube map is created to represent the diffuse lighting in a low resolution, akin to the filtered specular probe.
It shares the same optimizable parameters and is average-pooled to the mip level with $r=1$ roughness.
The pre-filtering only involves the first term in Eq.~\ref{supp:euq:render}.

%% file: supp/sections/2-implementation_details.tex
\section{Implementation Details}

\label{supp:implementation_details}

\noindent \textbf{SDF network}. We parametrize the SDF field with an MLP to increase surface water-tight and smoothness.
We choose the MLP architecture from \cite{mildenhallNeRFRepresentingScenes2020}, which consists of 6 frequency bands for positional encoding, and 8 linear layers, each having 256 neurons, followed by ReLU activations.
We implicitly regularize the smoothness by increasing the Lipschitz property in the SDF field\cite{liuLearningSmoothNeural2022a}.

\noindent \textbf{Material network}. The material model is a small MLP with hash-encoding~\cite{mullerInstantNeuralGraphics2022} as the materials query is computationally extensive.
The MLP consists of two linear layers, each having 32 neurons, followed by ReLU activations.
The hash-encoding has a spatial resolution of 4096 and the rest configures are the same as~\cite{munkbergExtractingTriangular3D2022}.
To reduce computation, we predict all material channels at once with one backbone network.
Besides, we introduce inductive bias of materials of clothed humans in the real world,
by providing minimum and maximum values for each materials channel.
We follow~\cite{Zhang2021NeRFactorNF} to limit the albedo $\mathbf{k}_d \in [0.03, 0.8]$, and the roughness $\mathbf{k}_r \in [0.08, 1]$.
The texels in the environment light are randomly initialized between $[0.25, 0.75]$.

\noindent \textbf{Motion networks}. For the motion module, we use the same MLP architecture as~\cite{chenSNARFDifferentiableForward2021,wangARAHAnimatableVolume}, which is similar to our SDF MLP.
To resolve the problem where the training pose variation is too limited for skinning field learning (\eg, self-rotation video without any limbs movements), 
we initialize the MLP with the pre-trained skinning model provided by~\cite{wangARAHAnimatableVolume},
and impose $\ell_2$ norm for the skinning weights logits between our predictions and the ground truth from SMPL~\cite{loperSMPLSkinnedMultiperson2015}.
We ablate the design choices in Sec.~\ref{supp:additional_ablations:skinning}.
For the non-rigid modeling, we use another 4-layer ReLU MLP with a 4-frequency-band positional encoding.
We also progressively anneal its encoding for 5k iterations as~\cite{parkNerfiesDeformableNeural2021}.
The weights of the last layer are initialized with a uniform distribution $\mathcal{U}(-10^{-5}, 10^{-5})$, \ie initializing the non-rigid offsets to be close to zero and not interfering with the major optimizations of geometry and materials.

\noindent \textbf{Optimization}. 
We use Adam~\cite{kingmaAdamMethodStochastic2017} as our default optimizer.
We optimize the subject for 5k steps for 1024$\times$1024 images or 10k steps 512$\times$512 images.
We disable the perturbed normal map during optimization as it leads to SDF collapsing abruptly at a certain step (\ie, all SDF values are positive or negative where marching tetrahedra fails).
The optimization process takes about an hour on a single NVIDIA GTX3090 GPU.
The indicative results with plausible quality appear after a few minutes, which is quite faster than our counterparts~\cite{pengNeuralBodyImplicit2021,pengAnimatableNeuralRadiance2021,wangARAHAnimatableVolume,xuSurfaceAlignedNeuralRadiance2022a}.
Such superior efficiency could largely accelerate downstream applications.
The training visualization is presented in the supplemental video.

\noindent \textbf{Tetrahedra grids}. We start with a tetrahedra grid with $128 \times 128$ resolution, including 192k tetrahedra and 37k vertices.
Each tetrahedron can produce at most 2 triangles by marching tetrahedra algorithm~\cite{munkbergExtractingTriangular3D2022,shenDeepMarchingTetrahedra2021,gaoLearningDeformableTetrahedral2020}.
To increase the resolution of the tetrahedra grid, we subdivide the grid at the 500th step.
To avoid the out-of-memory problem caused by the vast amount of floating meshes in the void space at the beginning of training, we pre-train the SDF network to \textbf{match a visual hull} of humans in canonical space.
The hull could be derived from either the skeleton capsules or the SMPL~\cite{loperSMPLSkinnedMultiperson2015} mesh.
Note that we only pre-train for 500 iterations, which leads to \textbf{a very coarse shape} akin to the visual hull rather than the given ground truth mesh.
The initialized mesh is presented in the training visualization part of the supplemental video.

%% file: supp/sections/3-additional_ablations.tex
\section{Additional Ablations}

\label{supp:additional_ablations}

\begin{figure}[tbph]
\centering
\includegraphics[width=\linewidth]{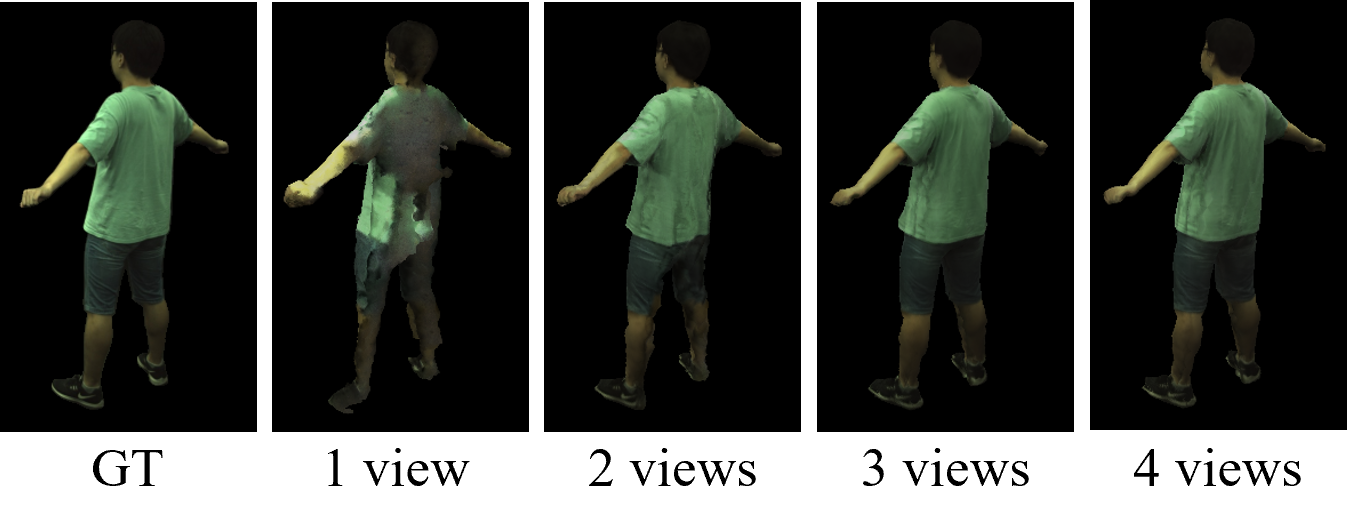}
\vspace{-1em}
\caption{\textbf{Ablation study of training views on the ZJU-MoCap 313 subject.}}
\label{fig:abla:view}
\end{figure}

\noindent \textbf{Number of Training Views}.
\label{supp:additional_ablations:num_views}
Table~\ref{table:abla:view} and Figure~\ref{fig:abla:view} show that giving one camera of view degrades the overall reconstruction quality, and multi-view consistency improves the final results.
The model can aggregate multi-view information for better shading optimization, thus leading to clearer surface materials.

\input{supp/tables/ablation-view}

\begin{figure}[tbph]
\centering
\includegraphics[width=\linewidth]{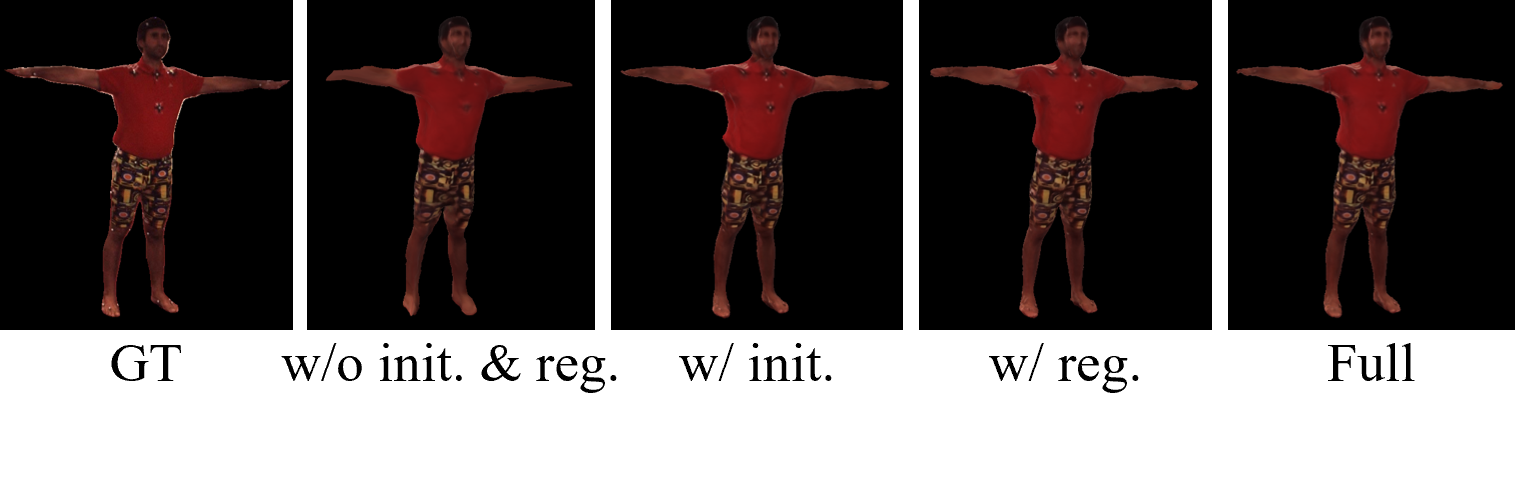}
\vspace{-1em}
\caption{\textbf{Ablation study of the skinning module on the H36M S9 subject.}}
\label{fig:abla:h36m}
\end{figure}

\begin{figure}[tbph]
\centering
\includegraphics[width=\linewidth]{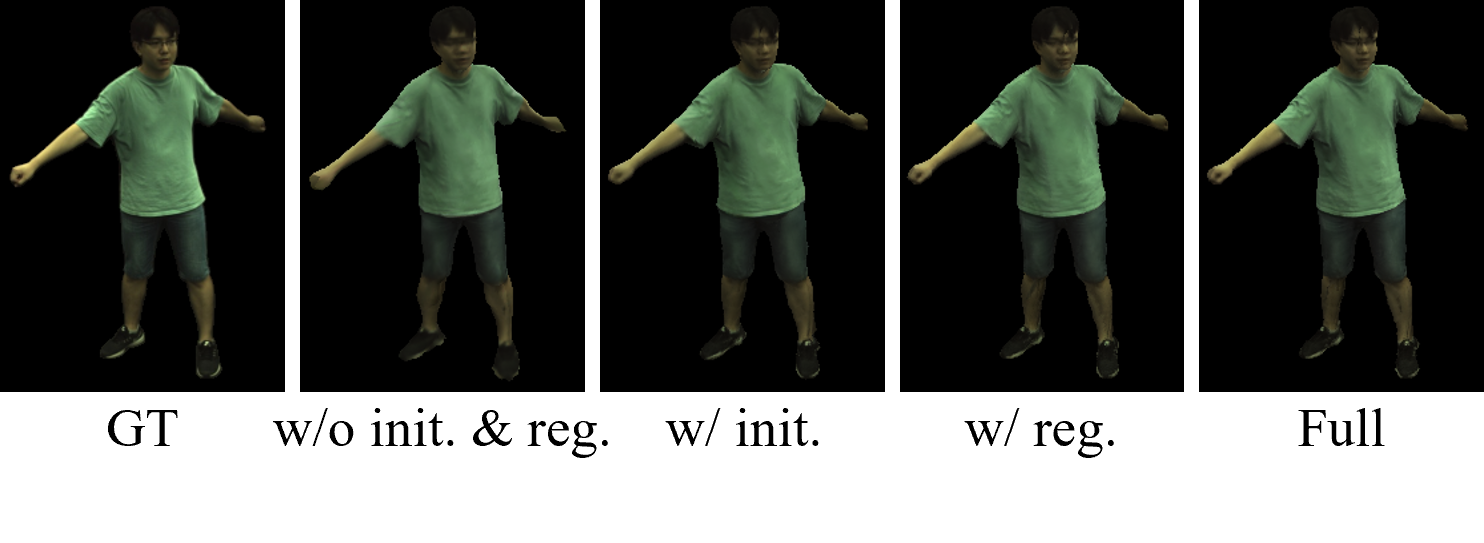}
\vspace{-1em}
\caption{\textbf{Ablation study of the skinning module on the ZJU-MoCap 313 subject.}}
\label{fig:abla:zju}
\end{figure}

\begin{figure}[tbph]
\centering
\includegraphics[width=\linewidth]{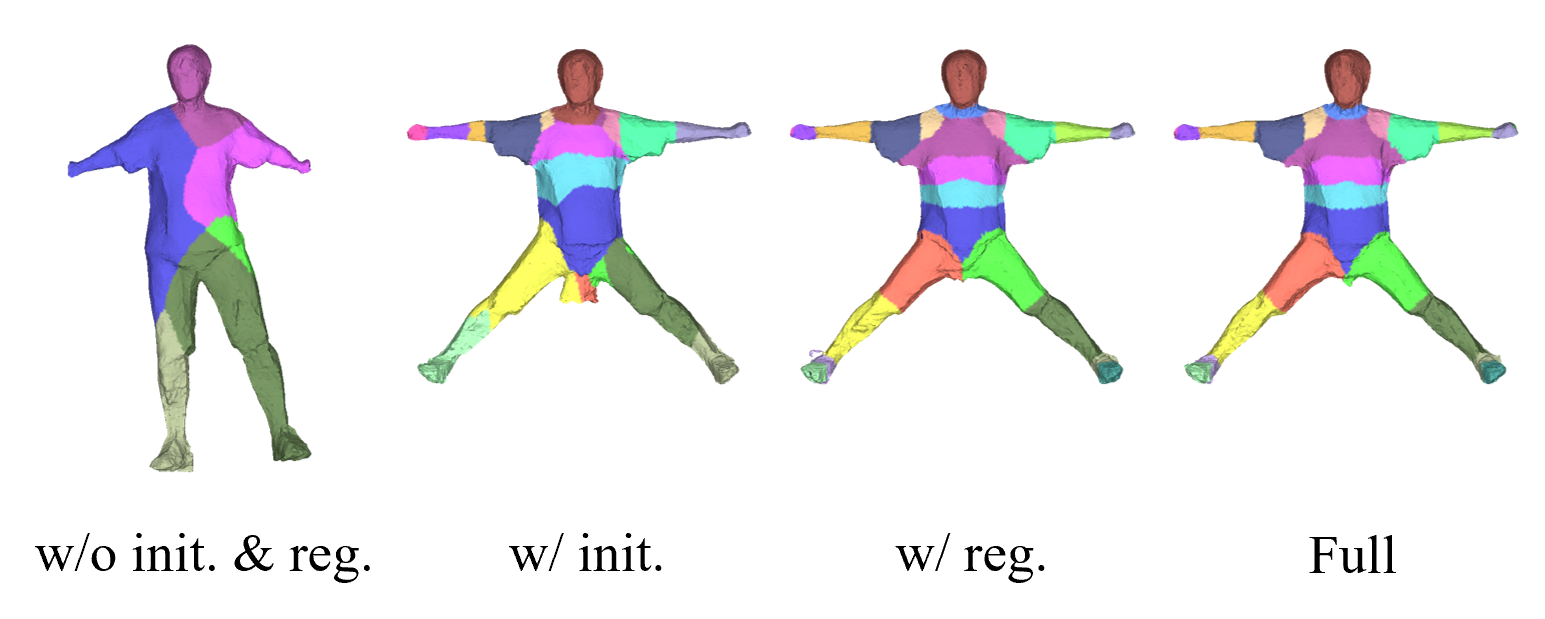}
\vspace{-1em}
\caption{\textbf{Ablation study of the skinning module on the ZJU-MoCap 313 subject.}}
\label{fig:abla:mesh:zju}
\end{figure}

\noindent \textbf{The effect of Skinning Module Design}
\label{supp:additional_ablations:skinning}
Table~\ref{table:ablation:skinning-h36m}-\ref{table:ablation:skinning-zju} and Figure~\ref{fig:abla:h36m}-\ref{fig:abla:mesh:zju} reveal that the initialization with pre-trained skinning net and the regularization on surface skinning improve the overall reconstruction quality.
The initialization provides skinning prior which helps to speed up geometry convergence. 
From Figure~\ref{fig:abla:h36m}-\ref{fig:abla:zju}, the geometry details improve with the initialization under the same training time.

The regularization on surface skinning prevents geometry degradation. 
Figure~\ref{fig:abla:mesh:zju} indicates that
our model can not learn correct canonical geometry without the initialization and the regularization.
The mesh distortion is reduced with the regularization.

\input{supp/tables/ablation-skinning}

\begin{figure}[tbph]
\centering
\includegraphics[width=\linewidth]{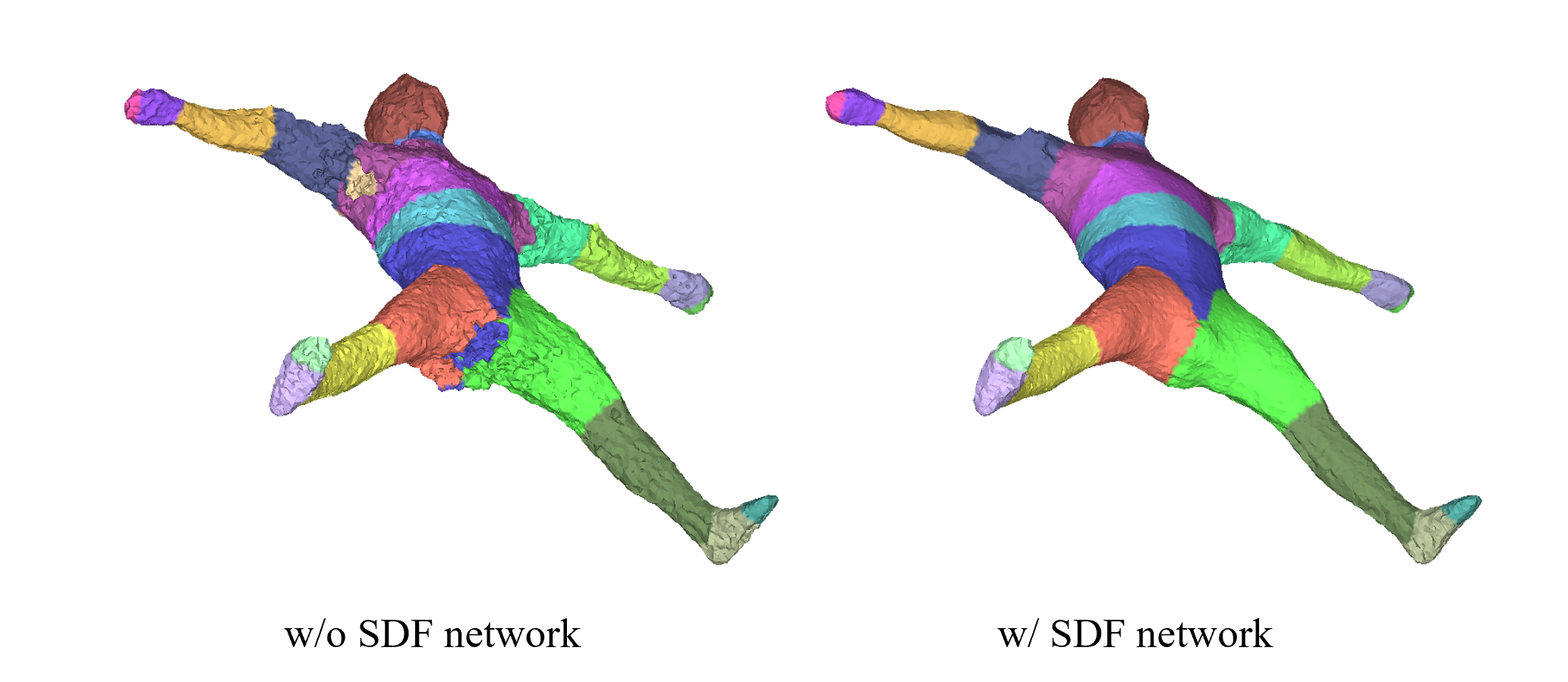}
\vspace{-1em}
\caption{\textbf{Ablation study of SDF field parametrization.}}
\label{fig:abla:sdf}
\end{figure}

\noindent \textbf{Effect of SDF Network}
\label{supp:additional_ablations:sdf}
The MLP parametrization of the SDF field keeps our surface both water-tight and smooth, as shown in Figure~\ref{fig:abla:sdf}.

%% file: supp/tables/ablation-view.tex
\begin{table}[htbp]
\centering
\caption{\textbf{Ablation results of training views on the ZJU-MoCap 313 subject}.}
\label{table:abla:view}
\begin{tabular}{ccccc}
\toprule
\multicolumn{1}{l|}{}    & \multicolumn{2}{c|}{Training pose}                         & \multicolumn{2}{c}{Novel pose}     \\ \hline
\multicolumn{1}{l|}{ZJU-MoCap 313}              & \multicolumn{1}{c|}{PSNR$\uparrow$}  & \multicolumn{1}{c|}{SSIM$\uparrow$}  & \multicolumn{1}{c|}{PSNR$\uparrow$}  & \multicolumn{1}{c}{SSIM$\uparrow$} \\ \midrule
\multicolumn{1}{l|}{1 view}   & \multicolumn{1}{l|}{24.39} & \multicolumn{1}{l|}{0.913} & \multicolumn{1}{l|}{21.45} & 0.869 \\
\multicolumn{1}{l|}{2 views}   & \multicolumn{1}{l|}{28.06} & \multicolumn{1}{l|}{0.945} & \multicolumn{1}{l|}{22.81} & 0.888 \\
\multicolumn{1}{l|}{3 views}   & \multicolumn{1}{l|}{28.50} & \multicolumn{1}{l|}{0.956} & \multicolumn{1}{l|}{23.17} & 0.894 \\
\multicolumn{1}{l|}{4 views}   & \multicolumn{1}{l|}{\textbf{29.04}} & \multicolumn{1}{l|}{\textbf{0.961}} & \multicolumn{1}{l|}{\textbf{23.20}} & \textbf{0.896} \\
\bottomrule
\end{tabular}
\end{table}

%% file: supp/tables/ablation-skinning.tex
\begin{table}[htbp]
\centering
\caption{
\textbf{The ablation on skinning module of H36M S9 dataset}.}
\label{table:ablation:skinning-h36m}
\resizebox{\columnwidth}{!}{%
\begin{tabular}{l|cc|cc}
\toprule
              & \multicolumn{2}{c|}{Training Pose} & \multicolumn{2}{c}{Novel Pose}     \\ \hline
H36M S9       & \multicolumn{1}{c|}{PSNR}  & SSIM  & \multicolumn{1}{c|}{PSNR}  & SSIM  \\ \midrule
w/o skinning init. \& reg.   & \multicolumn{1}{c|}{24.88} & 0.905 & \multicolumn{1}{c|}{21.97} & 0.851 \\
w/ skinning intialization & \multicolumn{1}{c|}{26.28} & 0.926 & \multicolumn{1}{c|}{24.47} & 0.897 \\
w/ skinning regularization       & \multicolumn{1}{c|}{26.24} & 0.925 & \multicolumn{1}{c|}{24.34} & 0.896 \\ \midrule
Full          & \multicolumn{1}{c|}{\textbf{26.29}} & \textbf{0.926} & \multicolumn{1}{c|}{\textbf{24.53}} & \textbf{0.899} \\
\bottomrule
\end{tabular}%
}
\end{table}

\begin{table}[htbp]
\centering
\caption{
\textbf{The ablation on skinning module of ZJU-MoCap 313 dataset}.}
\label{table:ablation:skinning-zju}
\resizebox{\columnwidth}{!}{%
\begin{tabular}{l|cc|cc}
\toprule
              & \multicolumn{2}{c|}{Training Pose} & \multicolumn{2}{c}{Novel Pose}     \\ \hline
ZJU-MoCap 313 & \multicolumn{1}{c|}{PSNR}  & SSIM  & \multicolumn{1}{c|}{PSNR}  & SSIM  \\ \midrule
w/o skinning init. \& reg.   & \multicolumn{1}{c|}{27.46} & 0.949 & \multicolumn{1}{c|}{20.31} & 0.831 \\
w/ skinning intialization & \multicolumn{1}{c|}{28.82} & 0.958 & \multicolumn{1}{c|}{23.08} & 0.893 \\
w/ skinning regularization       & \multicolumn{1}{c|}{28.80} & 0.959 & \multicolumn{1}{c|}{23.14} & 0.895 \\ \midrule
Full          & \multicolumn{1}{c|}{\textbf{29.05}} & \textbf{0.961} & \multicolumn{1}{c|}{\textbf{23.27}} & \textbf{0.897} \\
\bottomrule
\end{tabular}%
}
\end{table}

%% file: supp/sections/4-more_comparisons.tex
\section{More Comparisons}

\label{supp:more_comparisons}
\begin{figure*}[tbph]
\centering
\includegraphics[width=0.83\linewidth]{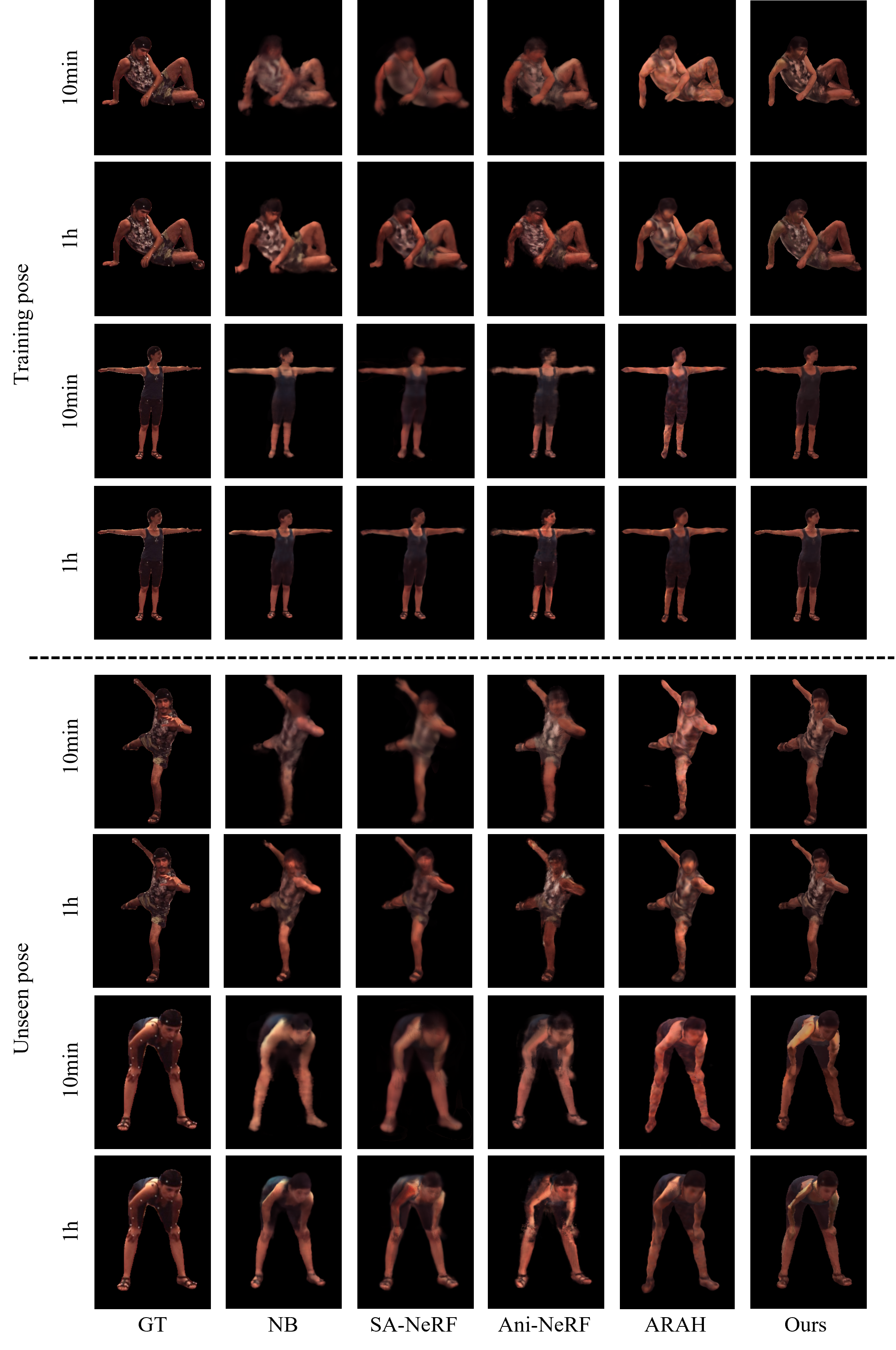}
\vspace{-1em}
\caption{\textbf{Qualitative results of novel pose synthesis on H36M dataset.} Zoom in for a better view.}
\label{fig:supp:h36m_qua}
\end{figure*}

\begin{figure*}[tbph]
\centering
\includegraphics[width=0.825\linewidth]{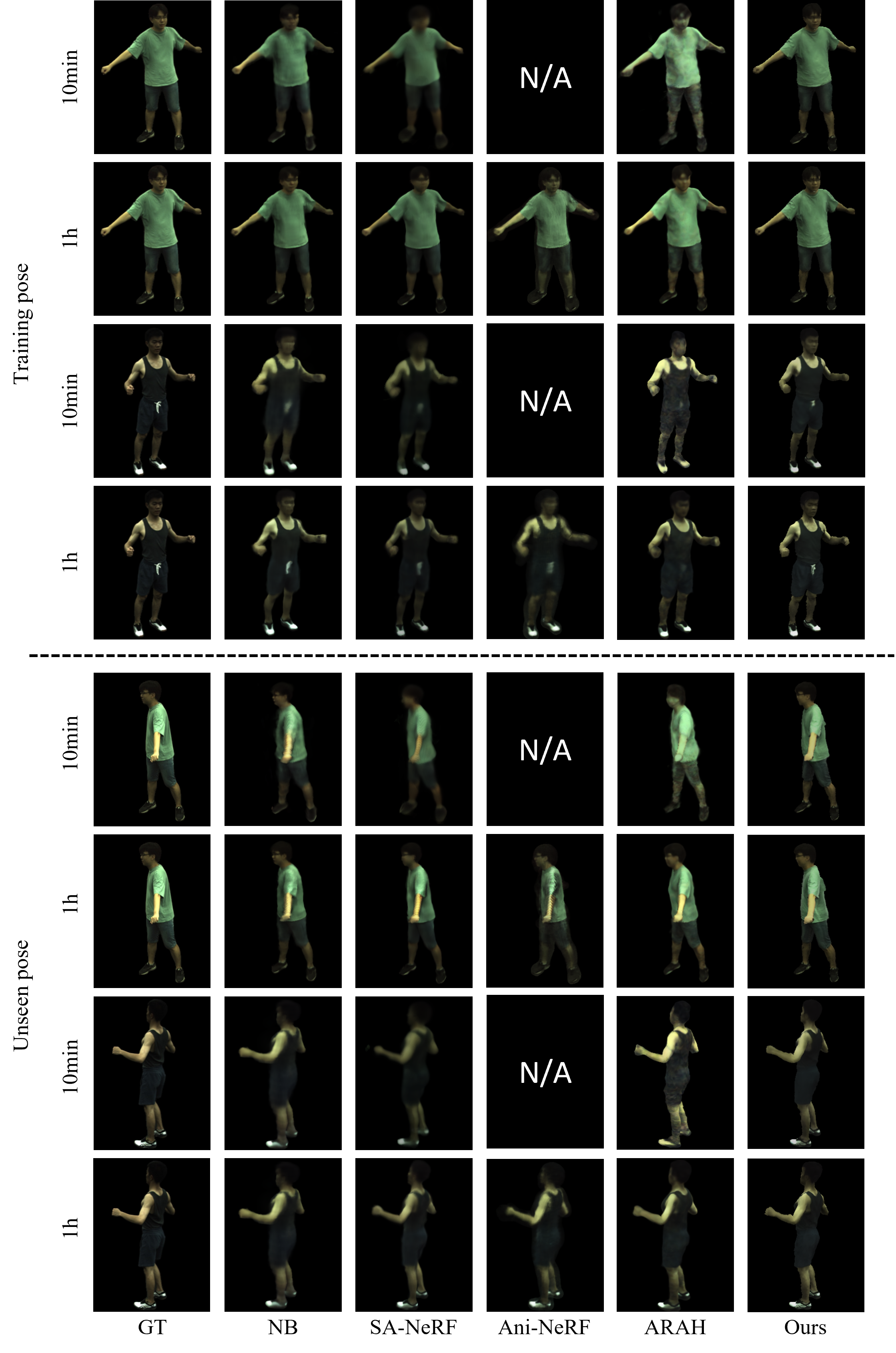}
\vspace{-1em}
\caption{\textbf{Qualitative results of novel pose synthesis on ZJU-MoCap dataset.} ``N/A'' denotes nothing to render due to no convergence. Zoom in for a better view.}
\label{fig:supp:zju_qua}
\end{figure*}

\begin{figure*}[tbph]
\centering
\includegraphics[width=0.8\linewidth]{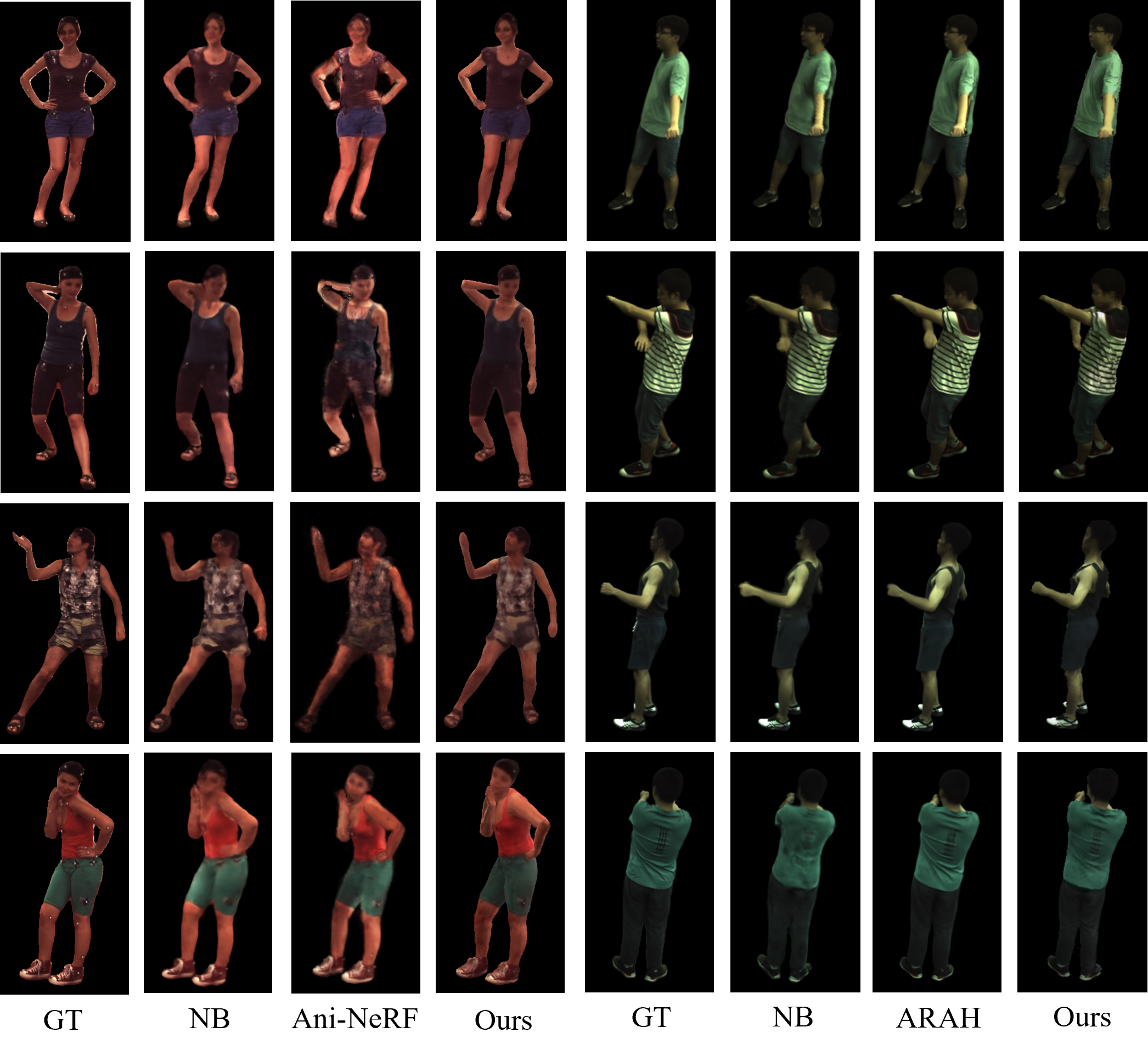}
\vspace{-1em}
\caption{\textbf{Qualitative results of novel pose synthesis on H36M and ZJU-MoCap datasets with the full models.} Zoom in for a better view.}
\label{fig:supp:full}
\end{figure*}

We present full quantitative comparisons in Table~\ref{tab:full:zju:trainingpose}, Table~\ref{tab:full:h36m:trainingpose}, Table~\ref{tab:full:zju:unseenpose}, and Table~\ref{tab:full:h36m:unseenpose}.
Meanwhile, more qualitative comparisons are illustrated in Figure~\ref{fig:supp:h36m_qua}, Figure~\ref{fig:supp:zju_qua}, and Figure~\ref{fig:supp:full}.

\input{supp/tables/full_table}

%% file: supp/tables/full_table.tex
\begin{table*}[thbp]
\centering
\caption{\textbf{Quantitative results of training pose novel view synthesis of H36M dataset}.}
\label{tab:full:h36m:trainingpose}
\resizebox{\textwidth}{!}{%
\begin{tabular}{l|cccccccccc}
\toprule
\multicolumn{1}{c|}{\multirow{3}{*}{}} & \multicolumn{10}{c}{Training pose} \\ \cline{2-11} 
\multicolumn{1}{c|}{} & \multicolumn{5}{c|}{PSNR} & \multicolumn{5}{c}{SSIM} \\ \cline{2-11} 
\multicolumn{1}{c|}{} & NB & SA-NeRF & Ani-NeRF & ARAH & \multicolumn{1}{c|}{Ours} & NB & SA-NeRF & Ani-NeRF & ARAH & Ours \\ \midrule
S1 & 22.87 & 23.71 & 22.05 & 24.45 & \multicolumn{1}{c|}{24.56} & 0.897 & 0.915 & 0.888 & 0.919 & 0.919 \\
S5 & 24.60 & 24.78 & 23.27 & 24.54 & \multicolumn{1}{c|}{24.51} & 0.917 & 0.909 & 0.892 & 0.918 & 0.920 \\
S6 & 22.82 & 23.22 & 21.13 & 24.61 & \multicolumn{1}{c|}{24.55} & 0.888 & 0.881 & 0.854 & 0.903 & 0.902 \\
S7 & 23.17 & 22.59 & 22.50 & 24.31 & \multicolumn{1}{c|}{24.05} & 0.914 & 0.905 & 0.890 & 0.919 & 0.916 \\
S8 & 21.72 & 24.55 & 22.75 & 24.02 & \multicolumn{1}{c|}{23.94} & 0.894 & 0.922 & 0.898 & 0.921 & 0.920 \\
S9 & 24.28 & 25.31 & 24.72 & 26.20 & \multicolumn{1}{c|}{25.99} & 0.910 & 0.913 & 0.908 & 0.924 & 0.919 \\
S11 & 23.70 & 25.83 & 24.55 & 25.43 & \multicolumn{1}{c|}{25.48} & 0.896 & 0.917 & 0.902 & 0.921 & 0.915 \\ \midrule
Average & 23.31 & 24.28 & 23.00 & 24.79 & \multicolumn{1}{c|}{24.72} & 0.902 & 0.909 & 0.890 & 0.918 & 0.916 \\ \bottomrule
\end{tabular}%
}
\end{table*}

\begin{table*}[thbp]
\centering
\caption{\textbf{Quantitative results of unseen pose novel view synthesis of H36M dataset}.}
\label{tab:full:h36m:unseenpose}
\resizebox{\textwidth}{!}{%
\begin{tabular}{l|cccccccccc}
\toprule
\multicolumn{1}{c|}{\multirow{3}{*}{}} & \multicolumn{10}{c}{Unseen pose} \\ \cline{2-11} 
\multicolumn{1}{c|}{} & \multicolumn{5}{c|}{PSNR} & \multicolumn{5}{c}{SSIM} \\ \cline{2-11} 
\multicolumn{1}{c|}{} & NB & SA-NeRF & Ani-NeRF & ARAH & \multicolumn{1}{c|}{Ours} & NB & SA-NeRF & Ani-NeRF & ARAH & Ours \\ \midrule
S1 & 21.93 & 22.67 & 19.96 & 23.08 & \multicolumn{1}{c|}{23.72} & 0.873 & 0.890 & 0.855 & 0.899 & 0.904 \\
S5 & 23.33 & 23.27 & 20.02 & 22.79 & \multicolumn{1}{c|}{23.13} & 0.893 & 0.881 & 0.840 & 0.890 & 0.898 \\
S6 & 23.26 & 23.23 & 23.64 & 24.04 & \multicolumn{1}{c|}{24.17} & 0.888 & 0.888 & 0.882 & 0.900 & 0.903 \\
S7 & 22.40 & 22.51 & 21.76 & 22.58 & \multicolumn{1}{c|}{22.72} & 0.888 & 0.898 & 0.869 & 0.891 & 0.889 \\
S8 & 20.78 & 23.06 & 21.63 & 22.34 & \multicolumn{1}{c|}{22.71} & 0.872 & 0.904 & 0.877 & 0.896 & 0.902 \\
S9 & 22.87 & 23.84 & 21.95 & 24.36 & \multicolumn{1}{c|}{24.54} & 0.880 & 0.889 & 0.871 & 0.894 & 0.895 \\
S11 & 23.54 & 24.19 & 22.55 & 24.78 & \multicolumn{1}{c|}{24.47} & 0.879 & 0.891 & 0.875 & 0.902 & 0.900 \\ \midrule
Average & 22.59 & 23.25 & 21.64 & 23.42 & \multicolumn{1}{c|}{23.64} & 0.882 & 0.892 & 0.867& 0.896 & 0.899 \\ \bottomrule
\end{tabular}%
}
\end{table*}

\begin{table*}[thbp]
\centering
\caption{\textbf{Quantitative results of training pose novel view synthesis of ZJU-MoCap dataset}.}
\label{tab:full:zju:trainingpose}
\resizebox{\textwidth}{!}{%
\begin{tabular}{l|cccccccccc}
\toprule
\multicolumn{1}{c|}{\multirow{3}{*}{}} & \multicolumn{10}{c}{Training pose} \\ \cline{2-11} 
\multicolumn{1}{c|}{} & \multicolumn{5}{c|}{PSNR} & \multicolumn{5}{c}{SSIM} \\ \cline{2-11} 
\multicolumn{1}{c|}{} & NB & SA-NeRF & Ani-NeRF & ARAH & \multicolumn{1}{c|}{Ours} & NB & SA-NeRF & Ani-NeRF & ARAH & Ours \\ \midrule
Twirl(313) & 30.56 & 31.32 & 29.80 & 31.60 & \multicolumn{1}{c|}{29.67} & 0.971 & 0.974 & 0.963 & 0.973 & 0.947 \\
Taichi(315) & 27.24 & 27.25 & 23.10 & 27.00 & \multicolumn{1}{c|}{24.21} & 0.962 & 0.962 & 0.917 & 0.965 & 0.919 \\
Swing1(392) & 29.44 & 29.29 & 28.00 & 29.50 & \multicolumn{1}{c|}{27.58} & 0.946 & 0.946 & 0.931 & 0.948 & 0.899 \\
Swing2(393) & 28.44 & 28.76 & 26.10 & 27.70 & \multicolumn{1}{c|}{25.91} & 0.940 & 0.941 & 0.916 & 0.940 & 0.890 \\
Swing3(394) & 27.58 & 27.50 & 27.50 & 28.90 & \multicolumn{1}{c|}{27.67} & 0.939 & 0.938 & 0.924 & 0.945 & 0.902 \\
Warmup(377) & 27.64 & 27.67 & 24.20 & 27.80 & \multicolumn{1}{c|}{26.69} & 0.951 & 0.954 & 0.925 & 0.956 & 0.926 \\
Punch1(386) & 28.60 & 28.81 & 25.60 & 29.20 & \multicolumn{1}{c|}{27.65} & 0.931 & 0.931 & 0.878 & 0.934 & 0.881 \\
Punch2(387) & 25.79 & 26.08 & 25.40 & 27.00 & \multicolumn{1}{c|}{25.68} & 0.928 & 0.929 & 0.926 & 0.945 & 0.908 \\
Kick(390) & 27.59 & 27.77 & 26.00 & 27.90 & \multicolumn{1}{c|}{24.08} & 0.926 & 0.927 & 0.912 & 0.929 & 0.840 \\ \midrule
Average & 28.10 & 26.19 & 28.27 & 28.51 & \multicolumn{1}{c|}{26.57} & 0.944 & 0.945 & 0.921 & 0.948 & 0.901 \\ \bottomrule
\end{tabular}%
}
\end{table*}

\begin{table*}[thbp]
\centering
\caption{\textbf{Quantitative results of unseen pose novel view synthesis of ZJU-MoCap dataset}.}
\label{tab:full:zju:unseenpose}
\resizebox{\textwidth}{!}{%
\begin{tabular}{l|cccccccccc}
\toprule
\multicolumn{1}{c|}{\multirow{3}{*}{}} & \multicolumn{10}{c}{Unseen pose} \\ \cline{2-11} 
\multicolumn{1}{c|}{} & \multicolumn{5}{c|}{PSNR} & \multicolumn{5}{c}{SSIM} \\ \cline{2-11} 
\multicolumn{1}{c|}{} & NB & SA-NeRF & Ani-NeRF & ARAH & \multicolumn{1}{c|}{Ours} & NB & SA-NeRF & Ani-NeRF & ARAH & Ours \\ \midrule
Twirl(313) & 23.95 & 24.33 & 22.80 & 24.40 & \multicolumn{1}{c|}{23.63} & 0.905 & 0.908 & 0.863 & 0.914 & 0.878 \\
Taichi(315) & 19.56 & 19.87 & 18.47 & 20.00 & \multicolumn{1}{c|}{20.42} & 0.852 & 0.863 & 0.795 & 0.881 & 0.850 \\
Swing1(392) & 25.76 & 26.27 & 18.44 & 26.20 & \multicolumn{1}{c|}{25.49} & 0.909 & 0.927 & 0.670 & 0.927 & 0.883 \\
Swing2(393) & 23.80 & 24.96 & 21.87 & 24.40 & \multicolumn{1}{c|}{24.31} & 0.878 & 0.900 & 0.836 & 0.915 & 0.883 \\
Swing3(394) & 23.25 & 24.24 & 17.69 & 25.20 & \multicolumn{1}{c|}{24.72} & 0.893 & 0.908 & 0.792 & 0.908 & 0.870 \\
Warmup(377) & 23.91 & 25.34 & 23.28 & 25.50 & \multicolumn{1}{c|}{24.80} & 0.909 & 0.928 & 0.901 & 0.933 & 0.894 \\
Punch1(386) & 25.68 & 27.30 & 25.55 & 27.00 & \multicolumn{1}{c|}{26.24} & 0.881 & 0.905 & 0.872 & 0.910 & 0.853 \\
Punch2(387) & 21.60 & 23.08 & 21.92 & 24.20 & \multicolumn{1}{c|}{24.06} & 0.870 & 0.890 & 0.838 & 0.917 & 0.889 \\
Kick(390) & 23.90 & 24.43 & 23.90 & 24.80 & \multicolumn{1}{c|}{25.79} & 0.870 & 0.889 & 0.887 & 0.896 & 0.873 \\ \midrule
Average & 23.49 & 24.42 & 21.55 & 24.63 & \multicolumn{1}{c|}{24.38} & 0.885 & 0.902 & 0.828 & 0.911 & 0.875 \\ \bottomrule
\end{tabular}%
}
\end{table*}

%% file: supp/sections/5-applications.tex
\section{Applications}
\label{supp:applications}
\begin{figure*}[tbph]
\centering
\includegraphics[width=1.0\linewidth]{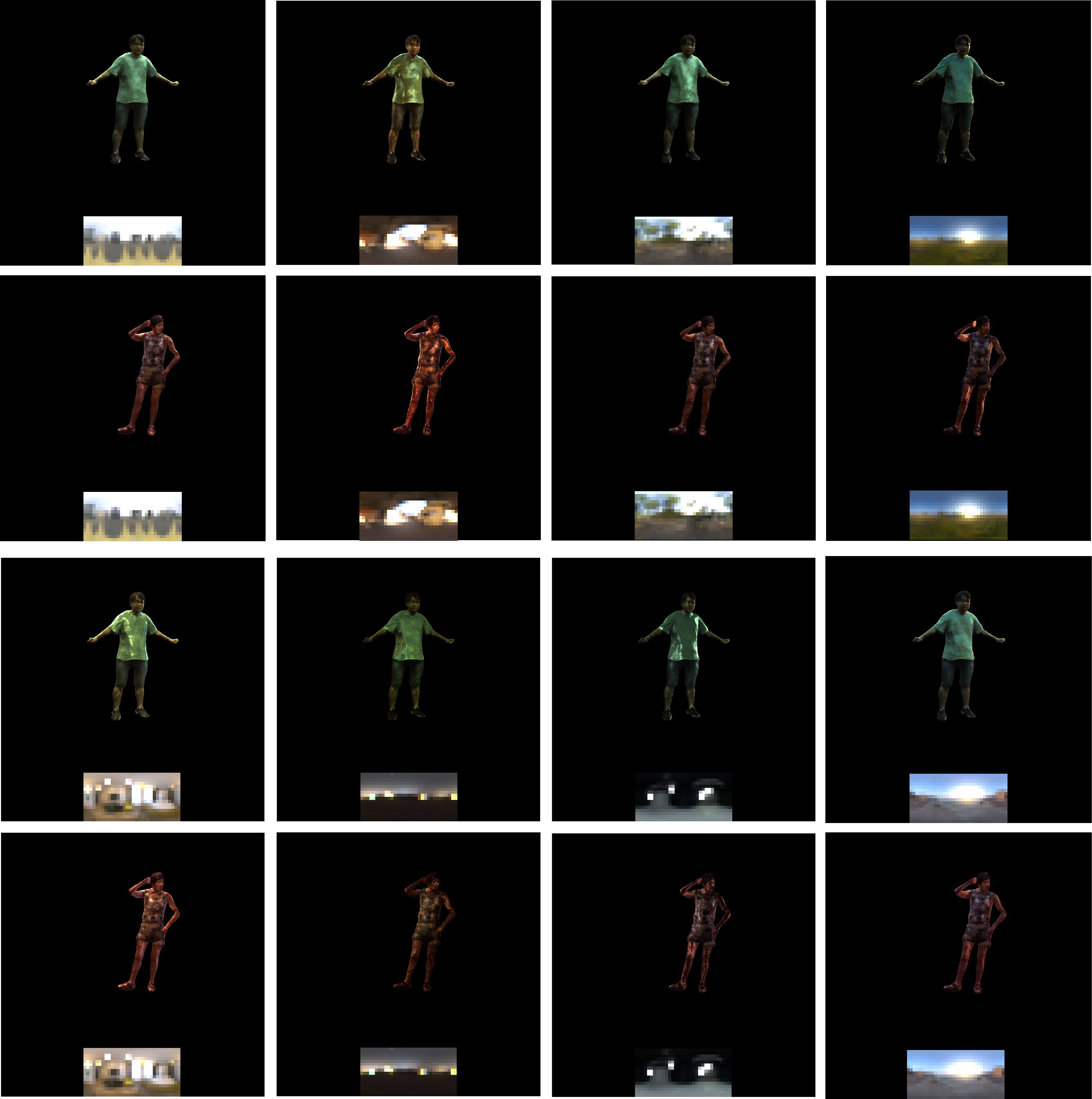}
\caption{\textbf{Relighting visualization.} Zoom in for a better view. We strongly encourage our readers to view the supplemental video for a more comprehensive visual perception.}
\label{fig:supp:relight}
\end{figure*}

\begin{figure*}[tbph]
\centering
\includegraphics[width=1.0\linewidth]{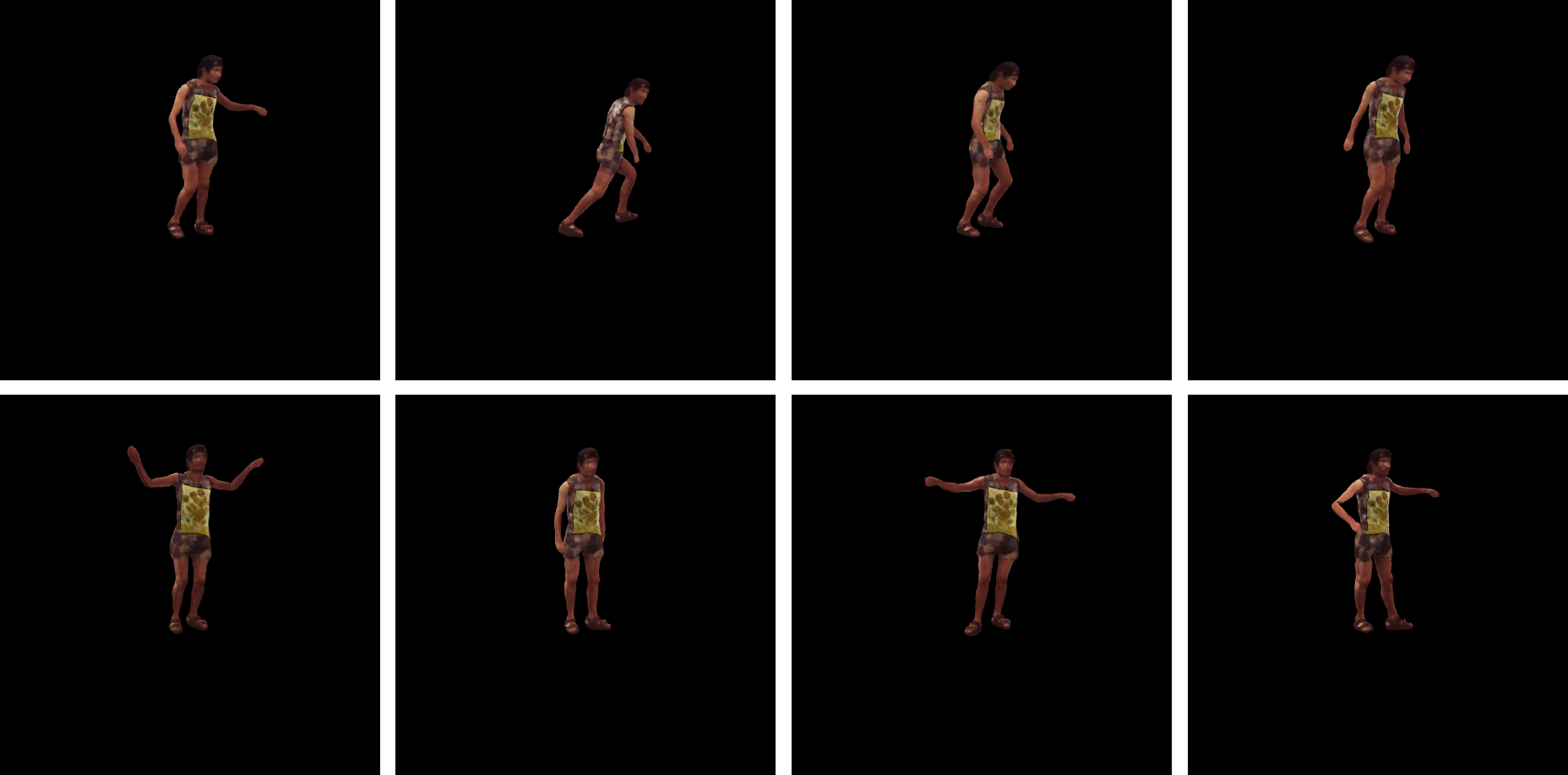}
\caption{\textbf{Texture editing visualization.} Zoom in for a better view. We strongly encourage our readers to view the supplemental video for a more comprehensive visual perception.}
\label{fig:supp:texture-edited}
\end{figure*}

We showcase \textbf{relighting}, \textbf{texture editing}, and \textbf{novel poses synthesis} on AIST dataset~\cite{Li2021AICM} in Figure~\ref{fig:supp:relight}, Figure~\ref{fig:supp:texture-edited}, and Figure~\ref{fig:supp:aist} separately. 
All the above applications are presented in the supplemental video.

\begin{figure*}[tbph]
\centering
\includegraphics[width=0.9\linewidth]{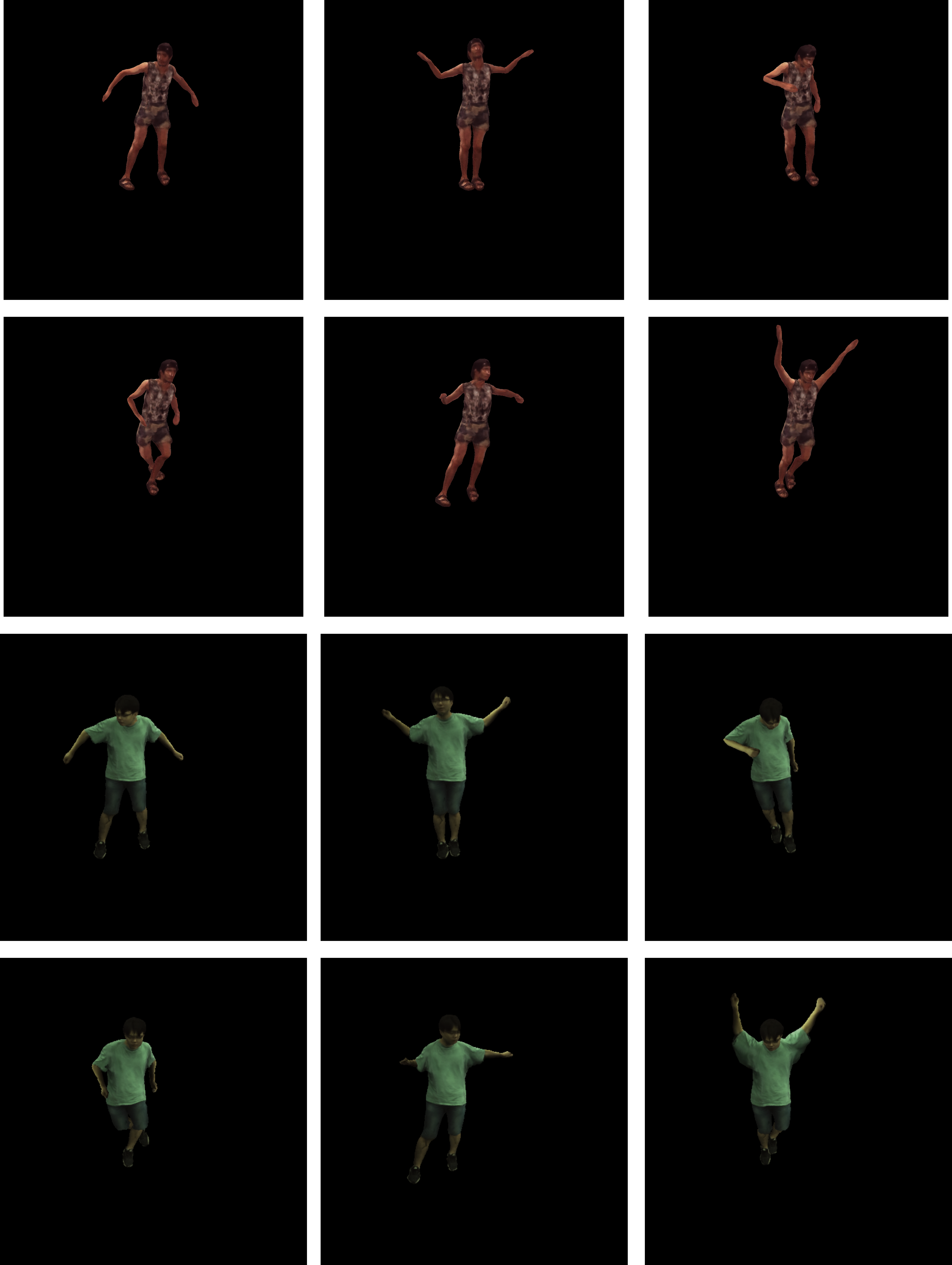}
\caption{\textbf{Extreme pose visualization.} Zoom in for a better view. We strongly encourage our readers to view the supplemental video for a more comprehensive visual perception.}
\label{fig:supp:aist}
\end{figure*}

%% file: supp/sections/6-mesh_evaluation.tex
\section{Mesh Visualizations}

\label{supp:mesh_evaluation}

\begin{figure*}
\centering
\includegraphics[width=0.7\linewidth]{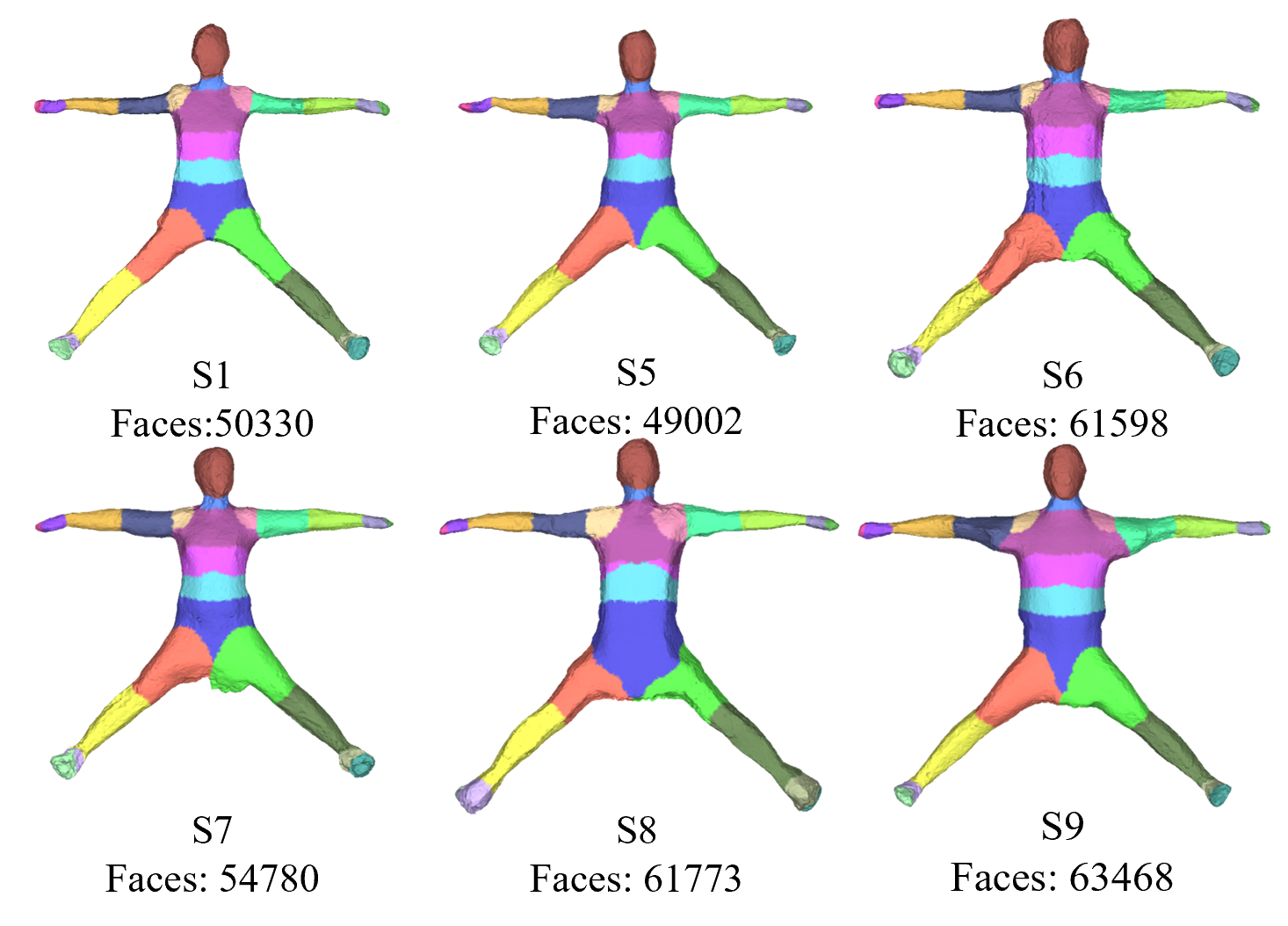}
\caption{\textbf{Mesh visualization on the H36M dataset.} Zoom in for a better view.}
\label{fig:supp:mesh:h36m}
\end{figure*}

We visualize the canonical mesh and present the number of faces of each mesh in Figure~\ref{fig:supp:mesh:h36m} and Figure~\ref{fig:abla:mesh:zju}.
Note that the number of faces for each mesh is quite small. 
Though increasing the resolution of tetrahedra grids may improve the details of both geometry and materials,
we do not conduct this experiment for it is orthogonal to our technical contributions.

\begin{figure*}
\centering
\includegraphics[width=0.7\linewidth]{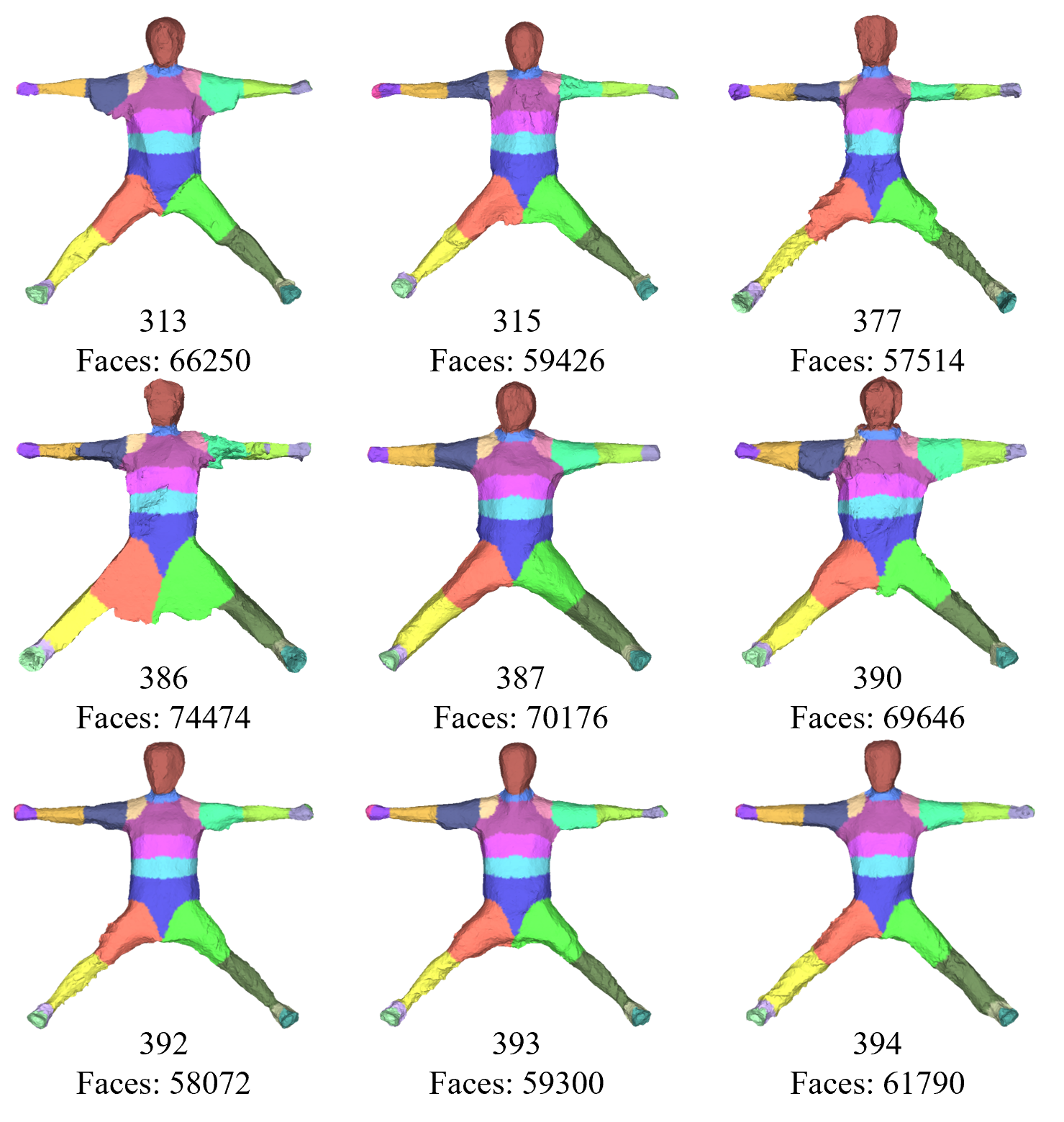}
\caption{\textbf{Mesh visualization on the ZJU-MoCap dataset.} Zoom in for a better view.}
\label{fig:supp:mesh:zju}
\end{figure*}

%% file: supp/sections/7-limitations_and_further_discussions.tex
\section{Limitations and Further Discussions}

\label{supp:limitation_discussions}

Our method is biased for shape-material ambiguity~\cite{wangARAHAnimatableVolume,liTAVATemplatefreeAnimatable2022,suANeRFArticulatedNeural2021,noguchiNeuralArticulatedRadiance2021,pengAnimatableNeuralRadiance2021,wangNeuSLearningNeural2021,zhangNeRFAnalyzingImproving2020}.
Taking subject 315 from ZJU-MoCap as an example, the strips in the T-shirt are modeled as ravines on the surface.
The high contrast color in the cloth surface makes our model biased for shape modeling.
That might be resolved by introducing additional surface regularizers or pre-defined parameters for the materials.

Need foreground mask to enable the mesh optimization, akin to shape-from-silhouette.
One future direction might be equipping our method with the ability to separate foreground and background automatically~\cite{jiangNeuManNeuralHuman2022,guoVid2Avatar3DAvatar2023b}.
It is also promising to model the background simultaneously during foreground subject optimization~\cite{jiangNeuManNeuralHuman2022,guoVid2Avatar3DAvatar2023b},
which eliminates the requirement of foreground mask processing.

Our method can digitize humans from visual footage, which may involve avatar misuse without the permission of the owners. 
Methods like implicit adversarial watermarks~\cite{chenFAWAFastAdversarial2021,liWatermarkingbasedDefenseAdversarial2021} that disable the neural nets inference could assist the video creation to protect their portrait rights.
Another concern is the deep fake misuse~\cite{nguyenDeepLearningDeepfakes2022}, which corrupts the identity in the visual footage rendered by our model.
Methods like deep fake detection~\cite{panDeepfakeDetectionDeep2020} could help to discover and prevent deep fake creations.
Besides, our method involves training with GPUs, which leads to carbon emissions and increasing global warming~\cite{pattersonCarbonEmissionsLarge2021}.